\documentclass[letterpaper,10pt,conference]{ieeeconf}
\IEEEoverridecommandlockouts

\usepackage{cite}
\usepackage{amsmath,amssymb,amsfonts}
\usepackage{algorithmic}
\usepackage{graphicx}
\usepackage{subcaption}
\usepackage{textcomp}
\usepackage{xcolor}
\usepackage{array}
\usepackage{hyperref}

\title{\LARGE \bf
Towards Assessing Compliant Robotic Grasping\\
from First-Object Perspective via Instrumented Objects
}
\author{Maceon Knopke$^{1}$, Liguo Zhu$^{1}$, Peter Corke$^{1}$, and Fangyi Zhang$^{1}$
\thanks{$^{1}$Authors are with the Queensland University of Technology (QUT) Centre for Robotics, 2 George Street, Brisbane City, 4000, Queensland, Australia. (email: fangyi.zhang@qut.edu.au)
        }
}

\begin{document}
\maketitle

\thispagestyle{empty}
\pagestyle{empty}

\begin{abstract}
Grasping compliant objects is difficult for robots -- applying too little force may cause the grasp to fail, while too much force may lead to object damage. A robot needs to apply the right amount of force to quickly and confidently grasp the objects so that it can perform the required task. Although some methods have been proposed to tackle this issue, performance assessment is still a problem for directly measuring object property changes and possible damage. To fill the gap, a new concept is introduced in this paper to assess compliant robotic grasping using instrumented objects. A proof-of-concept design is proposed to measure the force applied on a cuboid object from a first-object perspective. The design can detect multiple contact locations and applied forces on its surface by using multiple embedded 3D Hall sensors to detect deformation relative to embedded magnets. The contact estimation is achieved by interpreting the Hall-effect signals using neural networks. In comprehensive experiments, the design achieved good performance in estimating contacts from each single face of the cuboid and decent performance in detecting contacts from multiple faces when being used to evaluate grasping from a parallel jaw gripper, demonstrating the effectiveness of the design and the feasibility of the concept.
\end{abstract}

\section{Introduction}

Grasping and manipulating compliant objects such as fruit and fish are common tasks in daily human life, but it is still challenging for a robot to do so. Benefiting from the recent developments in computing hardware, vision and tactile sensors~\cite{navarro2023visuo}, and also machine learning techniques~\cite{dlinrobotics2018}, promising methods and solutions have been proposed for compliant object manipulation~\cite{delgado2015tactile,kaboli2016tactile,hogan2018tactile}.
However, there exists a common problem that good ways of directly measuring the stress or damage to the object are lacking~\cite{sanchez2018robotic,billard2019trends}, although vision-based or deformation-model-based methods can infer this and provide observations from outside the object.

\begin{figure}[tbp]
\centering
\includegraphics[width=1.0\linewidth]{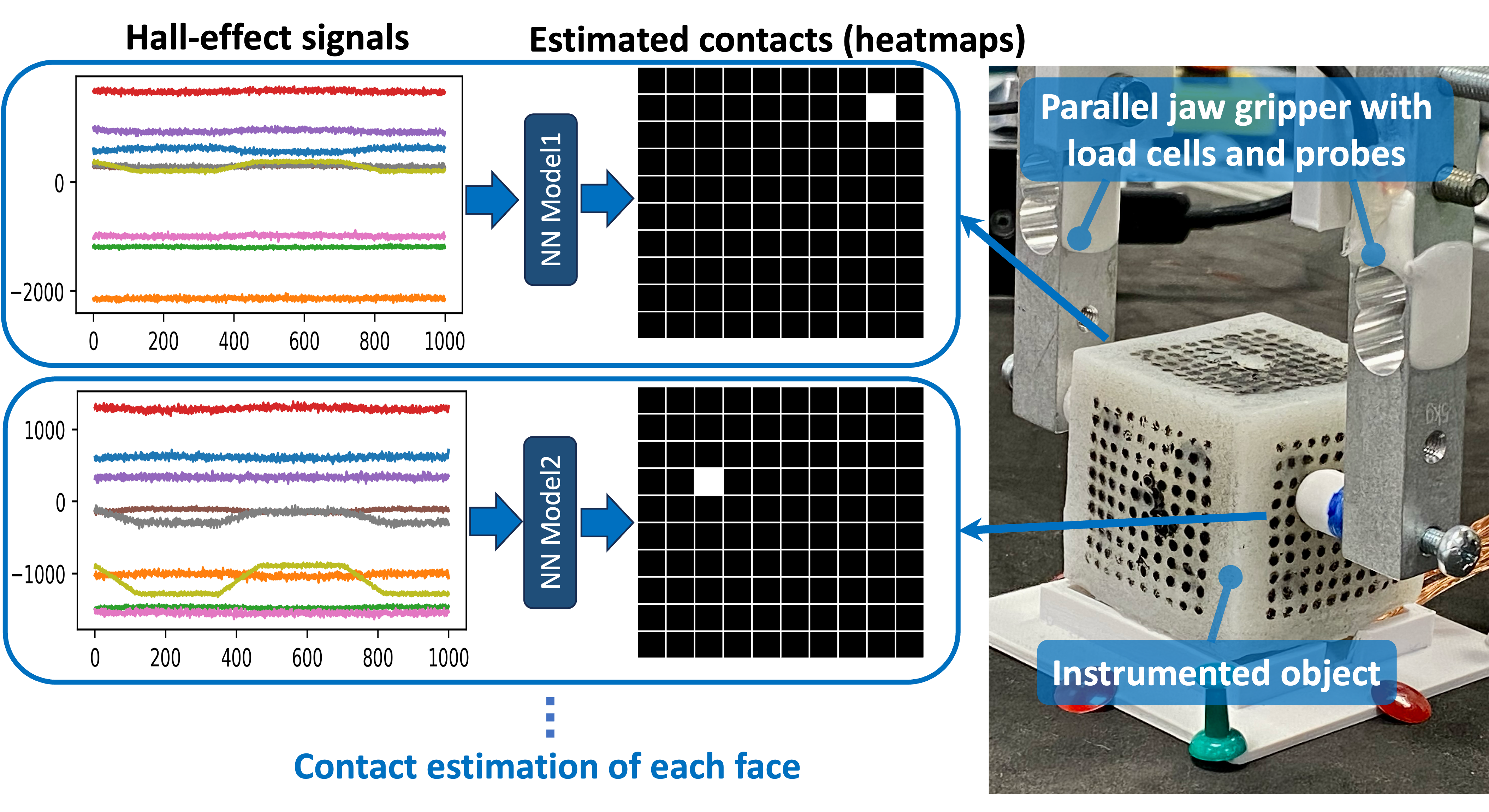}
\vspace{-8pt}
\caption{An instrumented object measuring robotic grasping from a parallel jaw gripper. Both contact location and normal force are detected in the form of heatmaps, by interpreting Hall-effect signals from the instrumented object using neural networks (NNs), one per face.
}
\label{fig:overall}
\vspace{-10pt}
\end{figure}

To fill this gap, and inspired by the recent advances in designing tactile sensors~\cite{chathuranga2016magnetic,9911183,10161344}, a new concept is introduced in this paper to assess compliant robotic grasping by adapting some of the tactile sensor designs to be used as mock objects (instrumented objects) for robots to grasp. As a proof of concept, the idea of using Hall-effect sensors to measure deformations~\cite{chathuranga2016magnetic} is extended in this paper to design a cuboid object that can detect contact locations and applied forces from a first-object perspective, as shown in Fig.~\ref{fig:overall}.
The design uses magnetometers, based on Hall-effect sensors, to observe the magnetic field created by a matrix of neodymium magnets embedded in the soft surface (silicone) of the object. As force is applied to the object, the magnetic field strength in each axis of the magnetometers changes. 
The contact estimation is then achieved by interpreting the Hall-effect signals using neural networks (NNs).

The design can accurately detect contact locations and applied forces on any face of the cuboid, with an average Euclidean location error of less than 1.27$\times10^{-2}$ pixels (each 26$\times$26 mm face has 10$\times$10 pixels), and an average force error of less than 3.23\% (0.479N). In the experiments with a parallel jaw gripper, where contacts are from multiple faces, the design can still accurately detect contact locations, although with relatively large force errors.
Nevertheless, these promising results demonstrate the effectiveness of the design and the feasibility of the new concept.
In particular, this paper makes three contributions:
\begin{itemize}
    \item a new concept is introduced to assess compliant robotic grasping from a first-object perspective using instrumented objects,
    \item a proof-of-concept design is proposed to detect contact locations and applied forces on a soft cuboid, which showed promising performance and demonstrated the feasibility of the new concept, and
    \item comprehensive experiments are conducted with the design, illustrating shortcomings of the current design and insights for future designs.
\end{itemize}

\section{Related Work}
To the best of our knowledge, there is no prior work in using instrumented objects to assess compliant robotic grasping from a first-object perspective. However, tactile sensing technologies, sensor designs, and their applications in robotics have been studied for decades~\cite{doi:10.1177/027836498900800301,LUO201754,xie2023learning}. 
The approach of using Hall-effect sensors to detect movement of a magnetic material embedded in a soft body has been adopted in a few papers for robotic tactile sensing~\cite{soft_3_axis,Vizzy,Spherical_Tactile_Sensors}. 
With Hall-effect sensors, a soft three-axis force sensor was designed for robot grippers to measure applied force and direction~\cite{soft_3_axis}.
In the design, a silicone hemisphere containing a neodymium magnet is secured to the base of the sensor containing a Hall-effect sensor. As the magnet is displaced, by applied normal and shear forces, the position is measured by the Hall-effect sensor and the force and direction can be estimated based on the silicone properties. Similarly, 3D Hall-effect sensors attached to flexible PCBs and covered by silicone coatings with magnets were mounted to a robot hand for detecting human-robot contacts~\cite{Vizzy}. A force sensor consisting of a hollow hemisphere with a magnet embedded at the top and four linear Hall-effect sensors in the base was also designed~\cite{Spherical_Tactile_Sensors}. With this design, multiple hemispheres can then be assembled in a 2D matrix to cover different parts of a robot such as robot hands, fingers or feet.
More recently, an automatic framework was proposed for designing soft sensing skins for arbitrary objects~\cite{10161344}, which could be extended to the implementation of instrumented objects. Similarly, various emerging vision-based tactile sensors~\cite{9911183} can also be adapted to use as instrumented objects. As an initial proof of concept, the Hall-effect-based approach is extended in this paper for instrumented objects.

\section{Instrumented Object Design}
\label{sec:instrumented_object_design}
To validate the new concept, and inspired by the soft force/tactile sensor designs~\cite{soft_3_axis,Vizzy,Spherical_Tactile_Sensors}, an instrumented object is designed to detect multiple contact points from multiple faces, by using multiple embedded 3D Hall sensors to observe deformation relative to embedded magnets. The contact estimation is achieved using a data-driven method.

\subsection{Hardware design}
\begin{figure}
    \centering
    \vspace{5pt}
    \begin{subfigure}{0.61\linewidth}
      \includegraphics[width=1.0\linewidth]{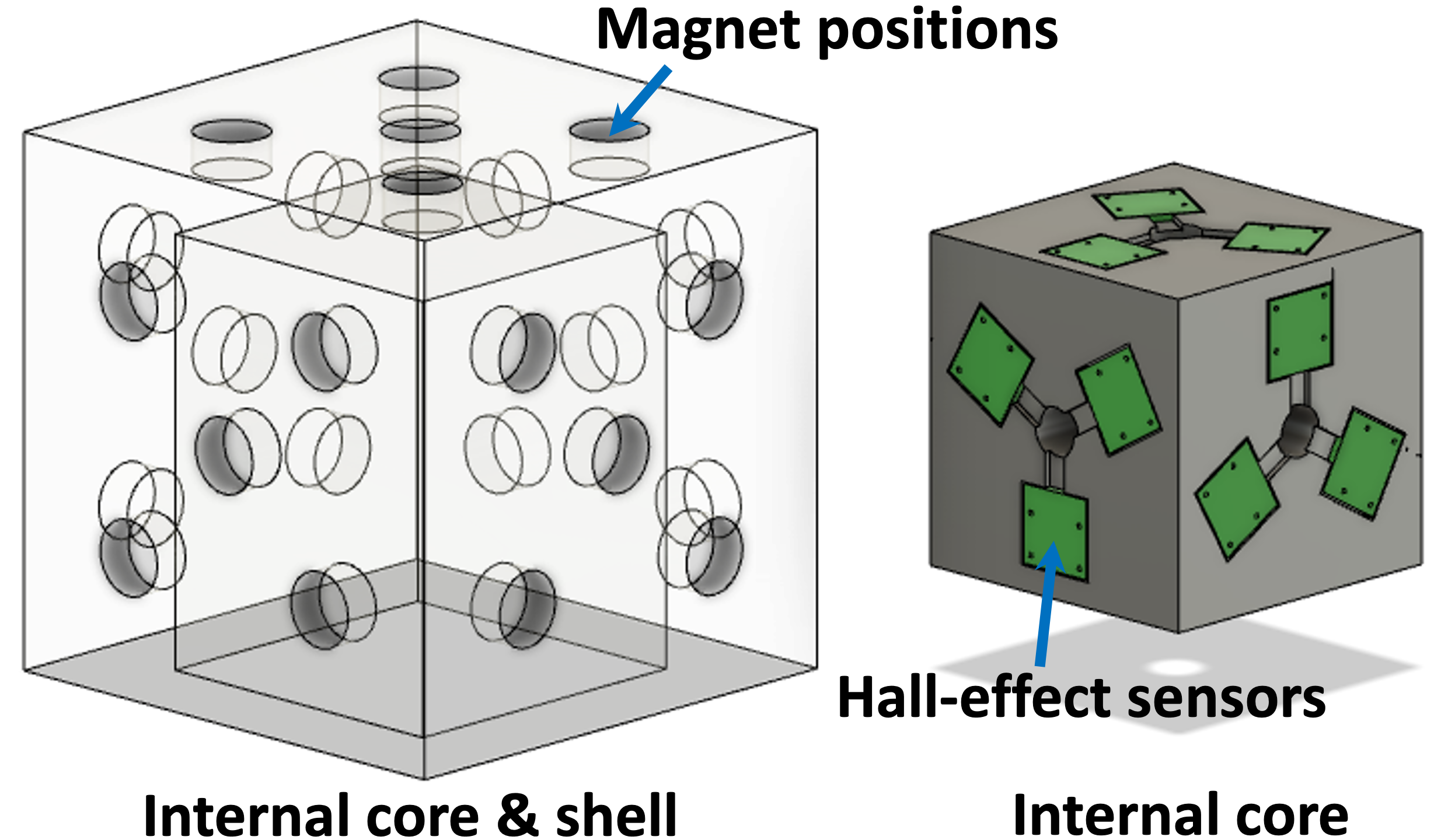}
      \caption{Overall design}
      \vspace{-2pt}
      \label{fig:overall_design}
      \end{subfigure}
      \begin{subfigure}{0.377\linewidth}
      \includegraphics[width=1.0\linewidth]{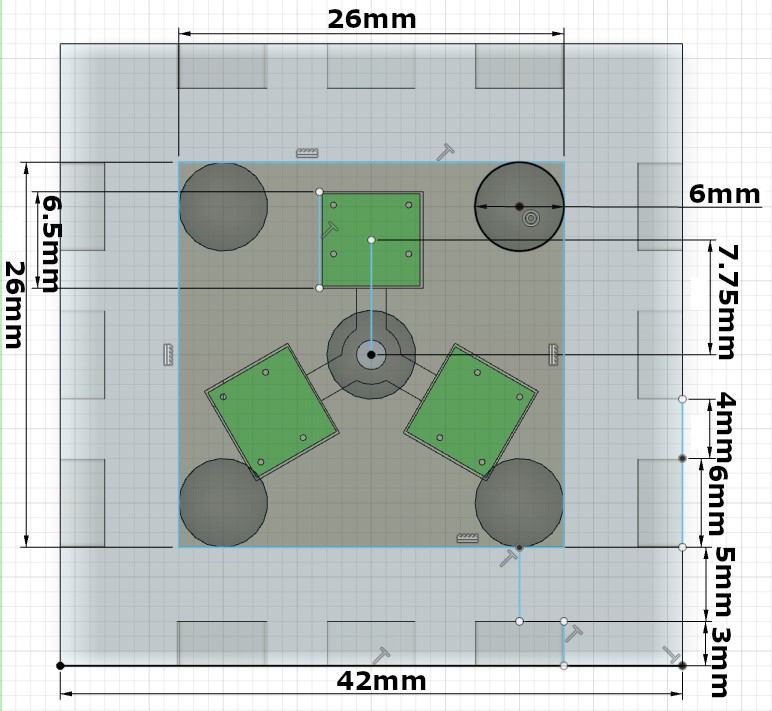}
      \caption{Dimension details}
      \vspace{-2pt}
      \label{fig:dimensions}
      \end{subfigure}
    \caption{Hardware design of the instrumented object.}
    \label{fig:InternalShell}
    \vspace{-15pt}
\end{figure}

The hardware part of the instrumented object consists of two parts: an internal core and a silicone shell, as shown in Fig.~\ref{fig:InternalShell}. The internal core is a solid cube with edges of length 26mm. The core has three 3D Hall sensors (TMAG5273A1)
on each face. All sensors are connected to an Arduino Mega 2560 Rev3 microcontroller
which samples the analog Hall signals. At least three independent measurements are needed to estimate three independent parameters (2D location and applied normal force) for each contact point~\cite{talcoth2011optimization}, which can be provided by a simple 3D Hall sensor. Therefore, three sensors are the minimum requirements for the detection of at most 3 contact points in parallel.
Although more sensors could help improve the robustness of the design, the required minimum number of Hall sensors are used in this design for initial proof of concept.

The other part of the instrumented object is the silicone shell which wraps the internal core. The silicone shell is a hollow cube with outer edges of length 42mm and inner edges of length 26mm. Each face of the shell has five N50 $6\times1$mm neodymium disk magnets embedded below its surface in five 3mm deep holes. 
A thin layer (about 2mm) of silicone is added over the magnets to seal them in place.  
The five magnets are arranged across each face as shown in Fig.~\ref{fig:InternalShell}. 
As the object is grasped, the silicone shell deforms and the strength of the magnetic field relative to each of the Hall sensors changes, therefore producing unique Hall-effect signals for contact estimation.

With this design, the shape of the internal sensor core and the silicone shell can be extensively changed to suit various situations and to satisfy the demands of assessing different robotic grasping or manipulation tasks, although a simple cube is used in this paper for proof of concept. 
The magnet type, Hall sensor type, and the relative locations between the magnets and Hall sensors are empirically determined in this proof of concept to minimize noise (reducing the influence of environmental magnetic fields including the Earth's magnetic field).
We do not claim its optimality. There is considerable scope for optimization in the types of sensors and magnets and their spatial arrangement.

\subsection{Contact estimation}
\label{sec:contact_estimation}
With the hardware design, the locations and force magnitudes of each contact are estimated using a learned model for each face. The model maps a Hall-signal frame $\mathbf{s} \in \mathbb{R}^{L}$ to a heatmap $\mathbf{h} \in \mathbb{R}^{M\times M}$: $\mathbf{h} = f_{\mathbf{\theta}}(\mathbf{s})$, 
where $\mathbf{\theta}$ represents the weights of the model; $L$ is the dimension of a Hall-signal frame, $L=9$ in this paper for the signals from the three Hall sensors (three for each sensor) for each face; and $M$ is the resolution of the 2D heatmap in one axis, $M=10$ in this paper. Only one-dimensional force magnitude (normal force) is considered in this paper for the proof of concept and also due to the limitation of the one-dimensional force sensor used for training data collection in Section\ref{sec:data_collection}, but the design can be extended to 3D force estimation by predicting 3-channel ($F_x$, $F_y$, $F_z$) heatmaps.

It is crucial to acknowledge that this scheme operates under the assumption that the Hall signal changes resulting from contacts on other faces are negligible, especially in cases involving multiple contacts from different faces. 
The scheme of independently estimating the contacts for each face can eliminate the need for additional training data encompassing extensive combinations of different contact points from different faces. Experiments in Section~\ref{sec:eval_real} showed the validity of this assumption with relatively small deformation from other faces, but non-trivial effects can be caused by large deformation from other faces and some remedies are discussed in Section~\ref{sec:design_considerations}.

\section{Fabrication}
\begin{figure}
    \centering
    \vspace{5pt}
    \begin{subfigure}{0.345\linewidth}
      \includegraphics[width=1.0\linewidth]{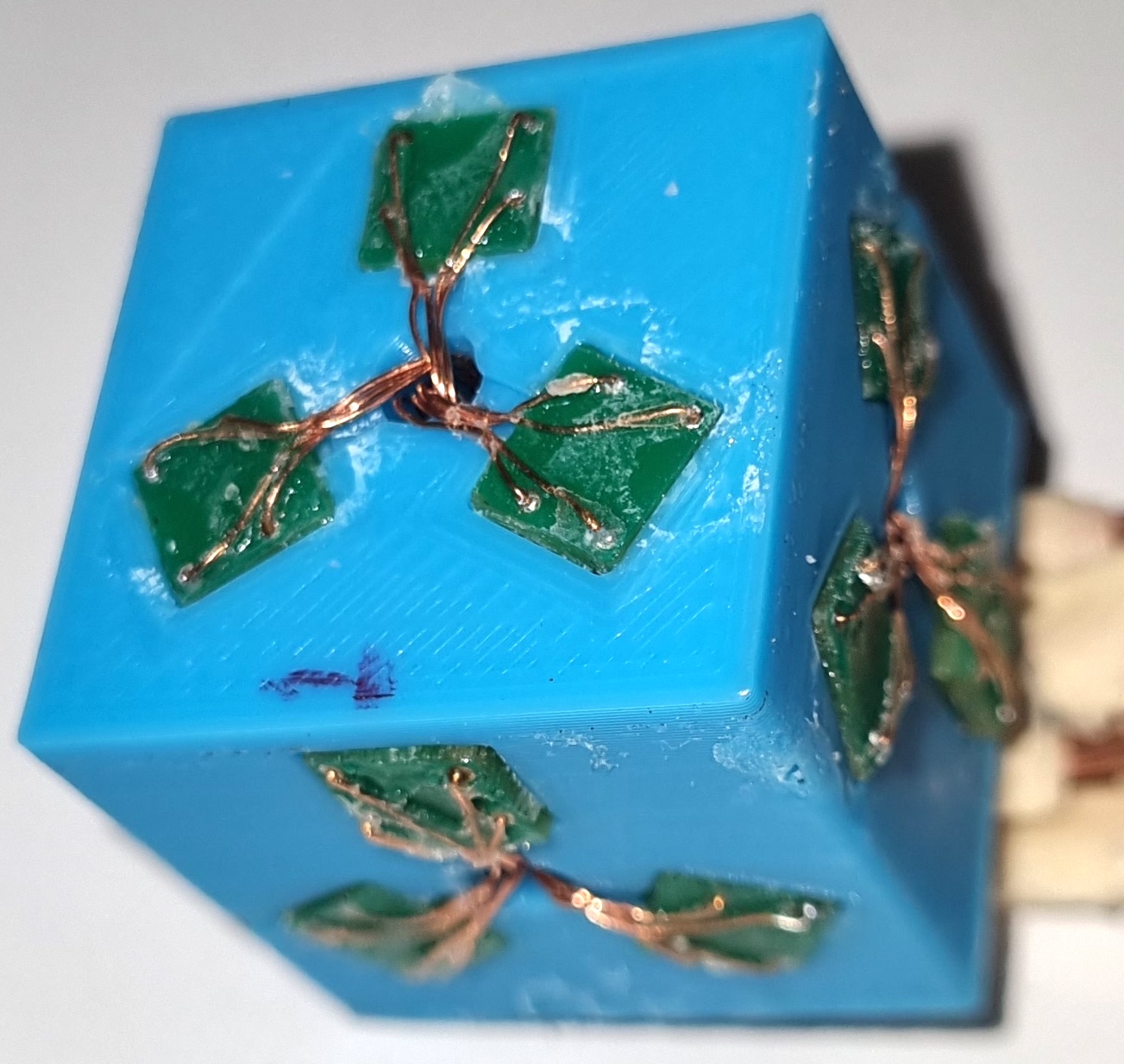}
      \caption{Internal core}
      \vspace{-2pt}
      \label{fig:real_object_internal}
    \end{subfigure}
    \begin{subfigure}{0.319\linewidth}
      \includegraphics[width=1.0\linewidth]{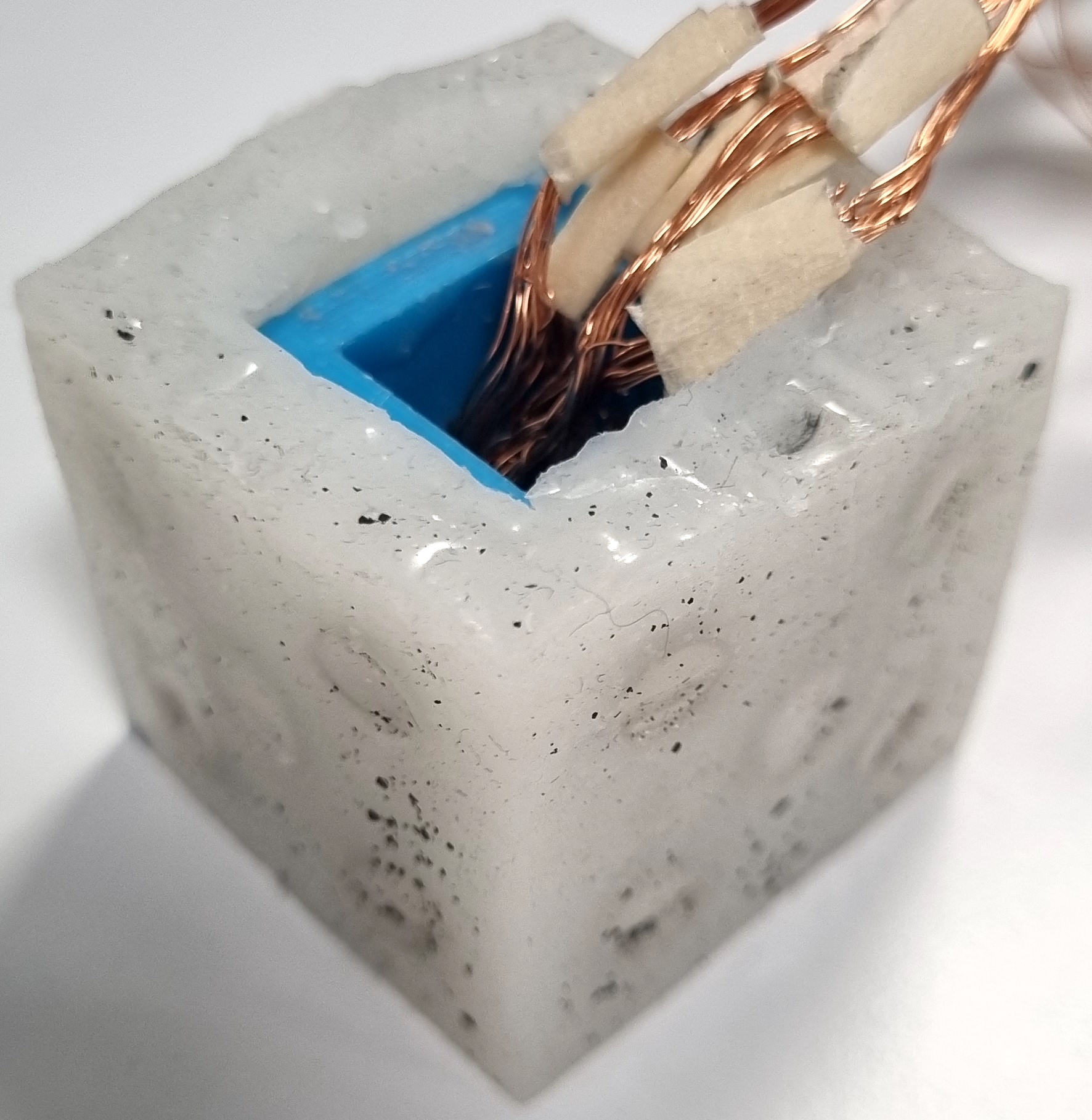}
      \caption{With a shell}
      \vspace{-2pt}
      \label{fig:real_object_with_shell}
    \end{subfigure}
    \begin{subfigure}{0.290\linewidth}
      \includegraphics[width=1.0\linewidth]{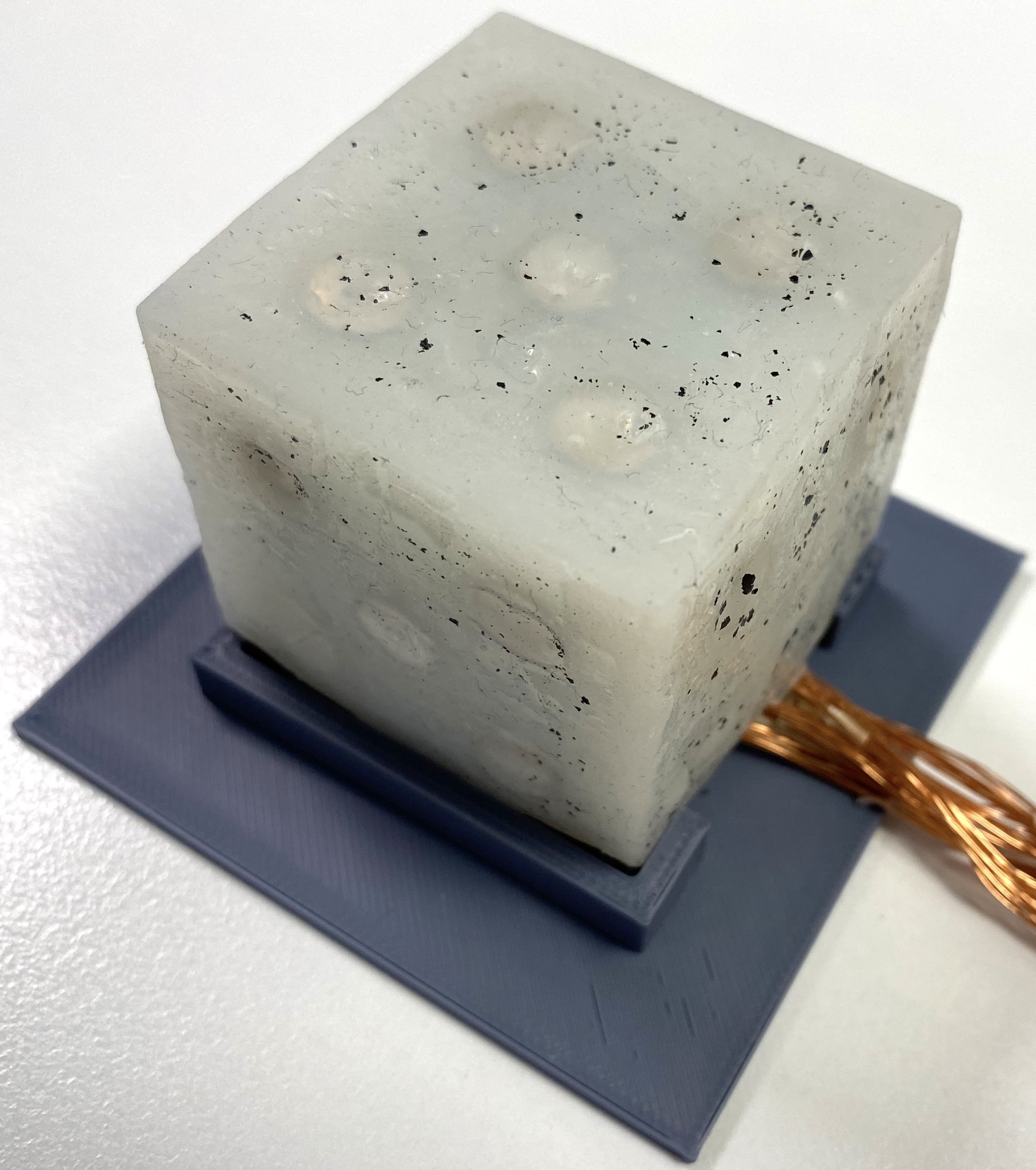}
      \caption{On a base}
          \vspace{-2pt}
      \label{fig:real_object_on_base}
    \end{subfigure}
    \caption{A fabricated instrumented object with a silicone shell covering the internal core.}
    \label{fig:real_instrumented_object}
    \vspace{-15pt}
\end{figure}

A prototype instrumented object is shown in Fig.~\ref{fig:real_instrumented_object}.
The first fabrication step is the assembly of the internal sensor core which is composed of a rigid cube (3D printed from PETG filament which has good layer adhesion and durability) and three TMAG5273 Hall sensors for each face. Each of the sensors is secured on a small custom PCB with a power decoupling capacitor, and is connected to the Arduino microcontroller's SDA, SCL and GND pins via thin 0.25mm enamel copper wires. For prototyping, an Arduino board outside the object was used, which prevented the use of one of the six faces -- 21 wires are required in total for connecting the sensors on the other five faces to the microcontroller and one face is used for passing these wires. In a complete implementation, an internal wireless microcontroller can be used together with a battery to eliminate the dependency on external wires.

The silicone shell was moulded using a 3D-printed mould. The internal core was pushed into the shell afterwards.
Platsil GEL-10 Prosthetic Grade Silicone was used in this paper to create the silicone shell with a desirable hardness of Shore A10. Different properties of the instrumented objects can be achieved by using different materials such as silicone with different levels of hardness.
To secure the magnets in place, the magnets were inserted into each of their moulded holes with the north poles facing outward. A thin layer of additional silicone was then spread over the surface to seal the magnets in place as in Fig.~\ref{fig:real_object_with_shell}.
All CAD models and source codes are open-sourced on the project page: 
\href{https://github.com/fangyizhang-x/inst-obj}{https://github.com/fangyizhang-x/inst-obj}.

\section{Experiments}
\label{sec:eval}

To verify the effectiveness of the design in measuring contact locations and force magnitudes when being grasped by a robotic gripper, extensive experiments are conducted in this section to evaluate its overall performance. We study how it performs with different contact locations,
and explore how the amount of training data influences its performance.

\subsection{Experimental setups}

The experiments were conducted by first collecting data from the prototype, and then using that data to train and test models for contact estimation.

\subsubsection{Data collection}
\label{sec:data_collection}

\begin{figure*}[thpb]
      \centering
      \vspace{5pt}
      \begin{subfigure}{0.19\linewidth}
      \includegraphics[width=1.0\linewidth]{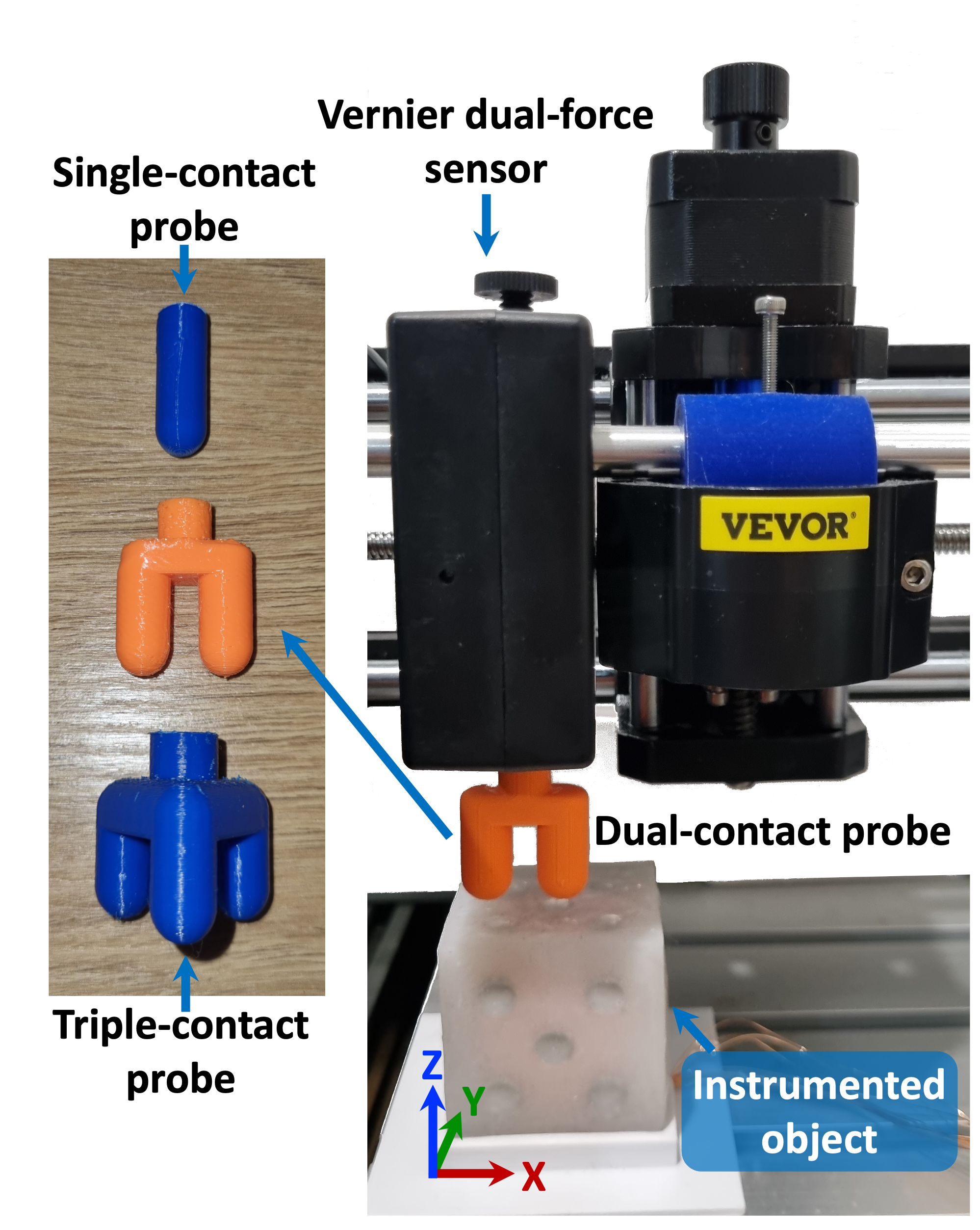}
      \caption{Data collection setups}
      \label{fig:Setup_Internal_shell}
      \vspace{-2pt}
      \end{subfigure}
      \begin{subfigure}{0.20\linewidth}
      \includegraphics[width=01.0\linewidth]{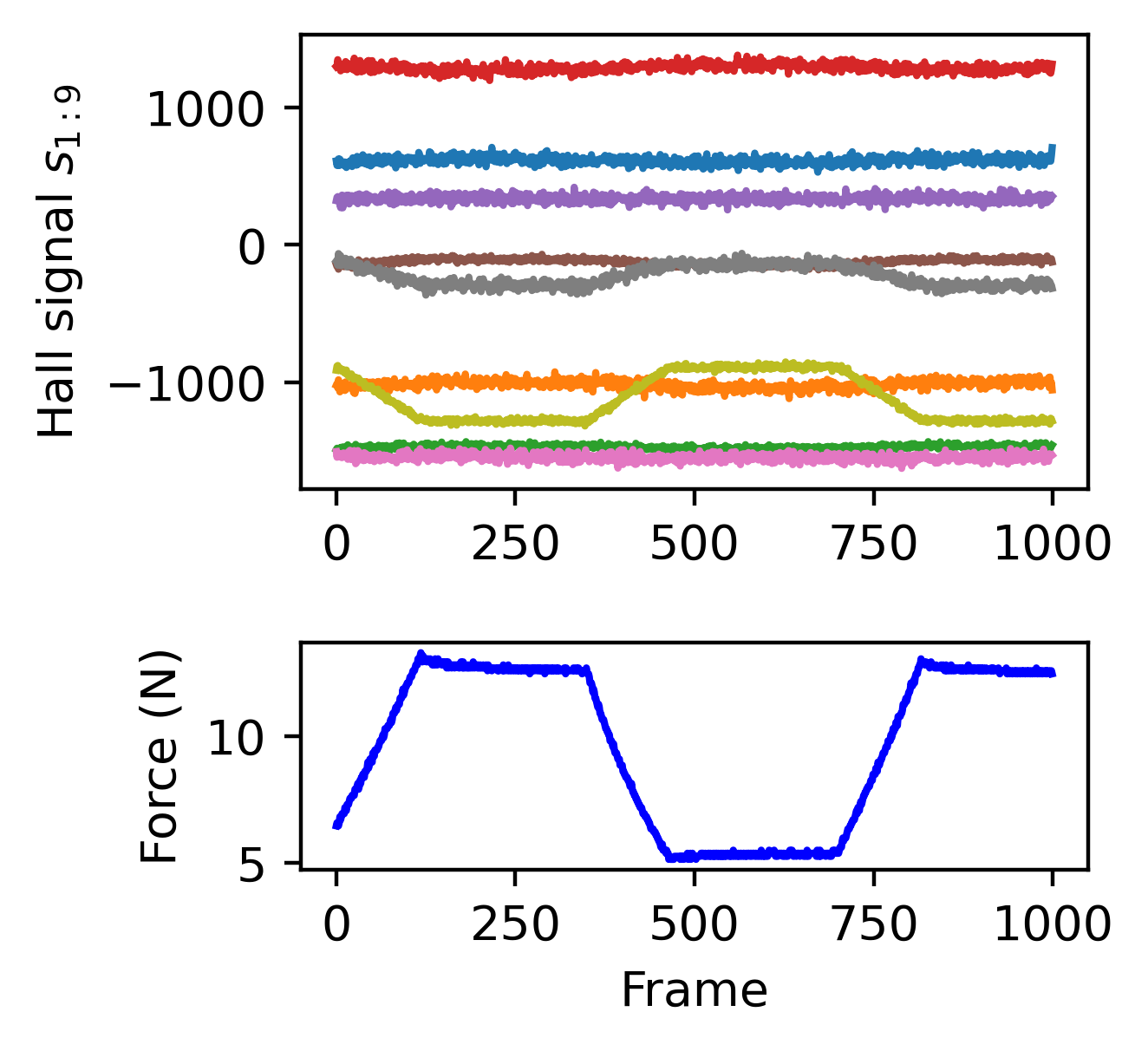}
      \caption{Force and Hall signals}
      \label{fig:raw_signal_vis}
      \vspace{-2pt}
      \end{subfigure}
      \begin{subfigure}{0.195\linewidth}
      \includegraphics[width=1.0\linewidth]{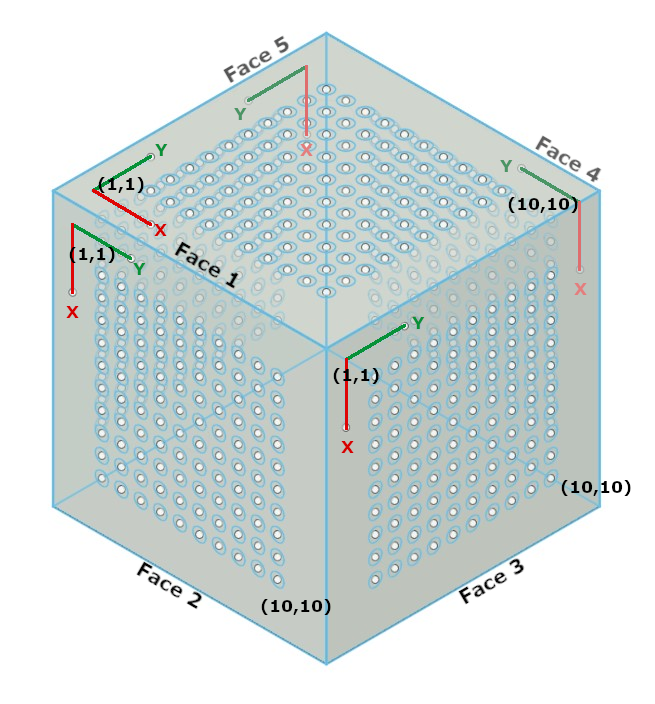}
      \caption{Faces and coordinates}
      \label{fig:face_definition}
      \vspace{-2pt}
      \end{subfigure}
      \begin{subfigure}{0.15\linewidth}
      \includegraphics[width=1.0\linewidth]{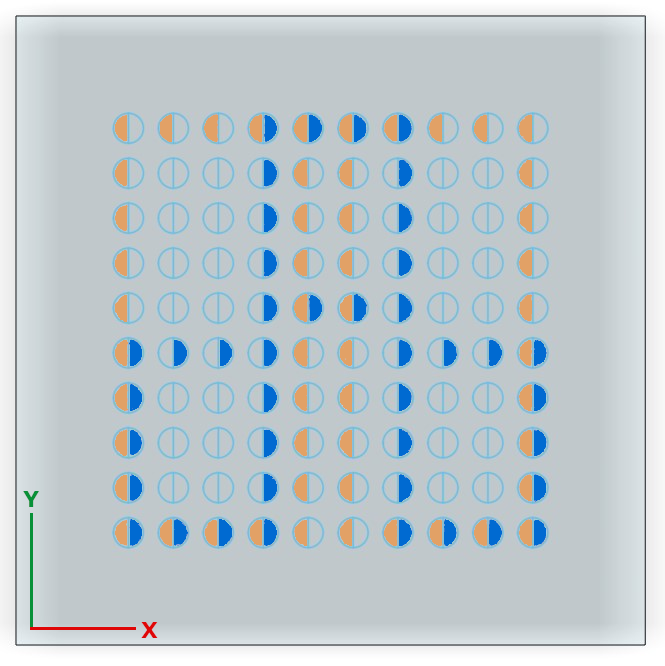}
      \vspace{0.1cm}
      \caption{Data coverage}
      \label{fig:data_coverage}
      \vspace{-2pt}
      \end{subfigure}
      \begin{subfigure}{0.24\linewidth}
      \includegraphics[width=1.0\linewidth]{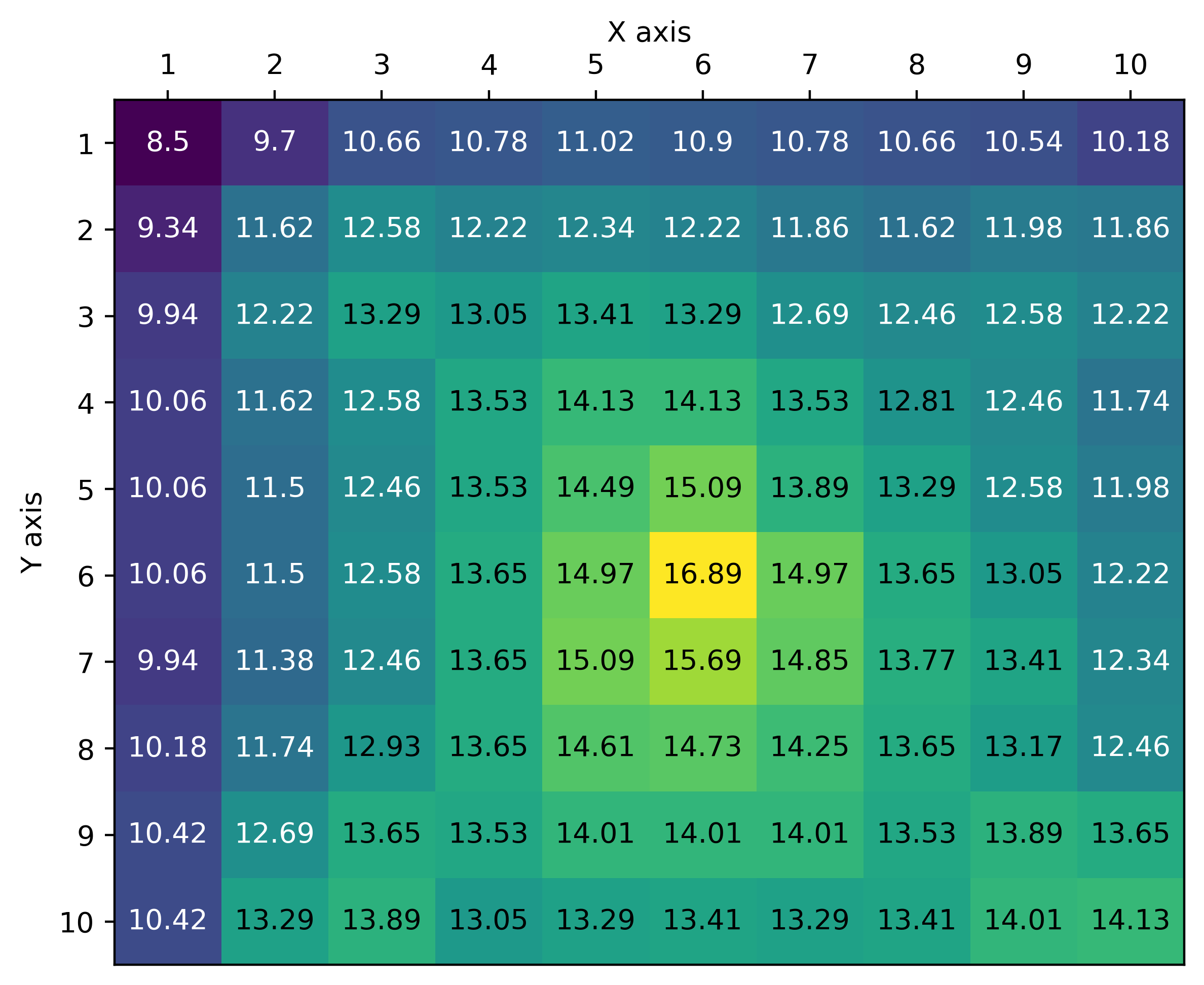}
      \vspace{1pt}
      \caption{Force ranges}
      \label{fig:loc_wise_force_range}
      \vspace{-2pt}
      \end{subfigure}
      \caption{Setups for data collection and data coverage. (a) A CNC machine positioned a Vernier force sensor and one of three types of probes (single-contact, dual-contact, and triple-contact) to collect data for the instrumented object. (b) One thousand Hall signal frames $\mathbf{s} \in \mathbb{R}^{9}$ and Vernier force measurements for one location (\textbf{Face1} (5,10)). (c) Face indexes and coordinates. (d) Data coverage for dual-contact (\textcolor{orange}{orange}) and triple-contact (\textcolor{blue}{blue}) cases. (e) \textbf{Face1} location-wise force ranges in Newton.}
      \label{fig:face_and_location_definition}
      \vspace{-15pt}
\end{figure*}

Data was collected for all five faces illustrated in Fig.~\ref{fig:face_definition}. When collecting the data a CNC (Computer Numerical Control) machine with a force sensor and 3D printed probe was used as shown in Fig.~\ref{fig:Setup_Internal_shell}. This allowed for accurate and repeatable placement of the probe in the ``X'' and ``Y'' axes (parallel to the face) along with a consistent movement in the ``Z'' axis when probing (normal to the face). During data collection for each face, the Hall-effect data $\mathbf{s} \in \mathbb{R}^{9}$ from the three 3D Hall sensors on that face was sampled along with force data from the probe, as shown in Fig.~\ref{fig:raw_signal_vis}. The force data was measured by a Vernier dual-range force sensor
set to a range of $\pm$ 50N, with its analog outputs sampled through the 10-bit ADC on the Arduino Mega which resulted in an accuracy of $\pm$ 0.12N.
During the data collection, the force and Hall-effect data were synchronized, and no smoothing was applied to the Hall-effect signals.

To evaluate the capability of the design in estimating multiple contact points, probes with one, two or three contact points were used, as shown in Fig.~\ref{fig:Setup_Internal_shell}. 
Location is defined for each face by a coordinate ($x$,$y$); $x,y \in [1,10]$, as shown in Fig.~\ref{fig:face_definition}.
All locations were covered for single-contact cases, the coverage of the cases with dual or triple contact points is encoded with colours as shown in Fig.~\ref{fig:data_coverage}. Due to setup limitations, the multiple contact points are correlated.

One thousand samples were collected for each contact location or multi-contact case, with z-axis displacement periodically changing between the minimum and maximum contact forces as shown Fig.~\ref{fig:raw_signal_vis}. The maximum force varies across the face (Fig.~\ref{fig:loc_wise_force_range}), as a consequence of local stiffness since we apply a constant z-axis displacement variation. 
Regarding the contact force of each point in multi-contact cases, it is assumed that the force measured by the Vernier force sensor ($F_{Vernier}$) is evenly distributed across the probe's contact points, i.e., each contact point has a force of $\frac{F_{Vernier}}{J}$ for a probe with $J$ contact points.

In total, there are 143k samples collected for each face $\mathbf{D}_{I}$: 100k for single-contact cases $\mathbf{D}^{1}_{I}$, 26k samples for dual-contact cases $\mathbf{D}^{2}_{I}$, 14k samples for triple-contact cases $\mathbf{D}^{3}_{I}$, and 3k samples for non-contact cases $\mathbf{D}^{0}_{I}$, where $I \in [1, \ldots, 5]$ is the face index.

\subsubsection{Model training}
\label{sec:training}
Following the machine learning protocol~\cite{bishop2006pattern}, the data collected for each face was randomly divided into three subsets: training set $\mathbf{D}_{I,\mbox{Tr}}$ (60\%, 85.8k samples), validation set $\mathbf{D}_{I,\mbox{Vl}}$ (20\%, 28.6k samples), and testing set $\mathbf{D}_{I,\mbox{Ts}}$ (20\%, 28.6k samples). This dataset division ratio applies to all contact cases (non-, single-, dual- or triple-contact cases, and any of their combinations).

A fully-connected (FC) neural network was adopted to model the projection from a Hall signal frame $\mathbf{s} \in \mathbb{R}^9$ to a contact heatmap $\mathbf{h} \in \mathbb{R}^{10\times 10}$. The force information was encoded into heatmaps with the force values in contact locations and zeros for the rest. The network consists of five FC layers with [600, 2000, 2000, 2000] units in the four hidden layers (with ReLU~\cite{fukushima1975cognitron} activation). This network architecture was used for prototyping purposes, but we do not claim its optimality for this contact estimation task. In training, both the inputs and outputs were normalized to [0,1], and a learning rate of 0.001 with a batch size of 2000 was used. Five networks were trained with the same settings for the five faces.

\begin{figure}[!t]
      \centering
      \begin{subfigure}{0.21\linewidth}
      \includegraphics[width=1.0\linewidth]{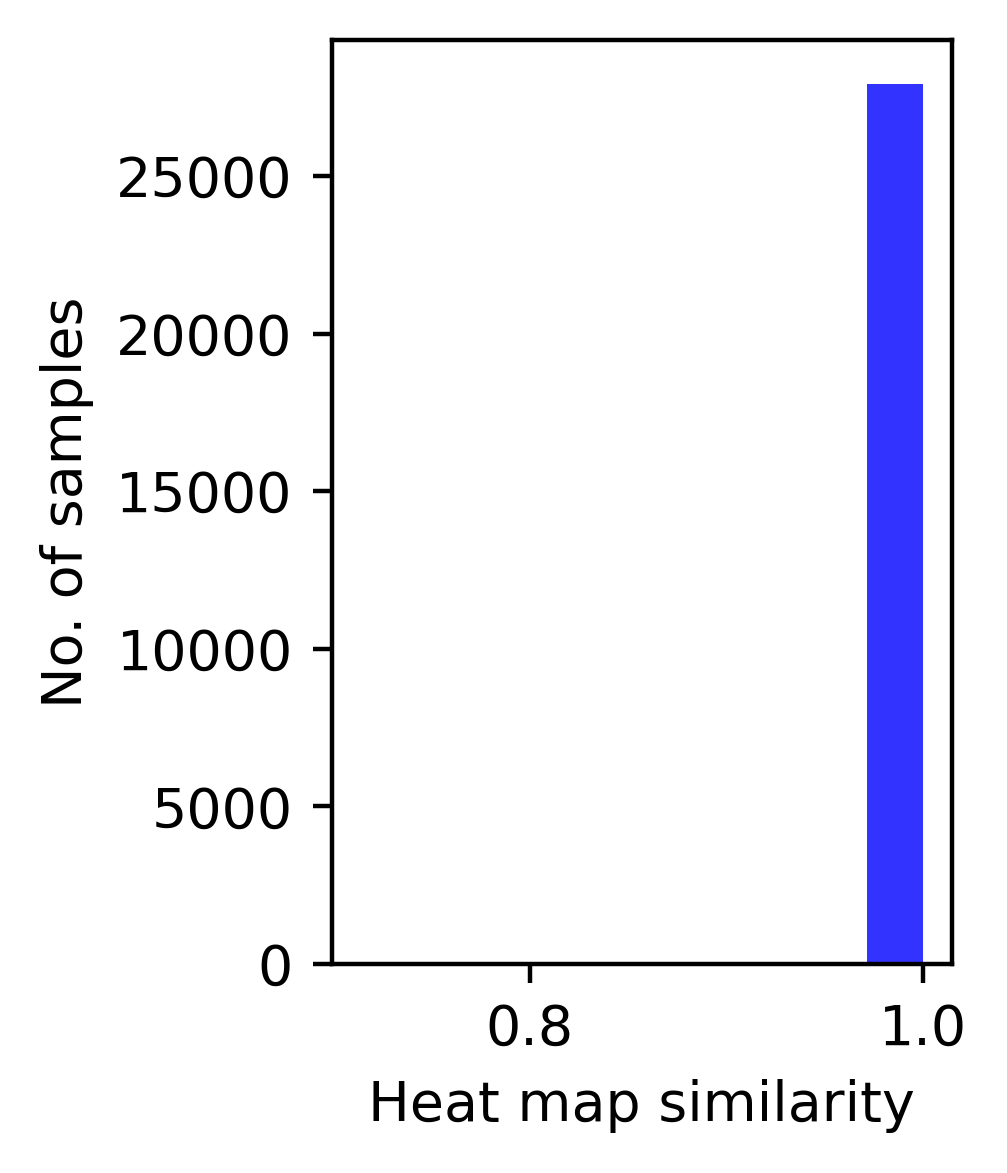}
      \caption{$A_{sim}$}
      \label{fig:hist_sim_face1}
      \vspace{-2pt}
      \end{subfigure}
      \begin{subfigure}{0.21\linewidth}
      \includegraphics[width=1.0\linewidth]{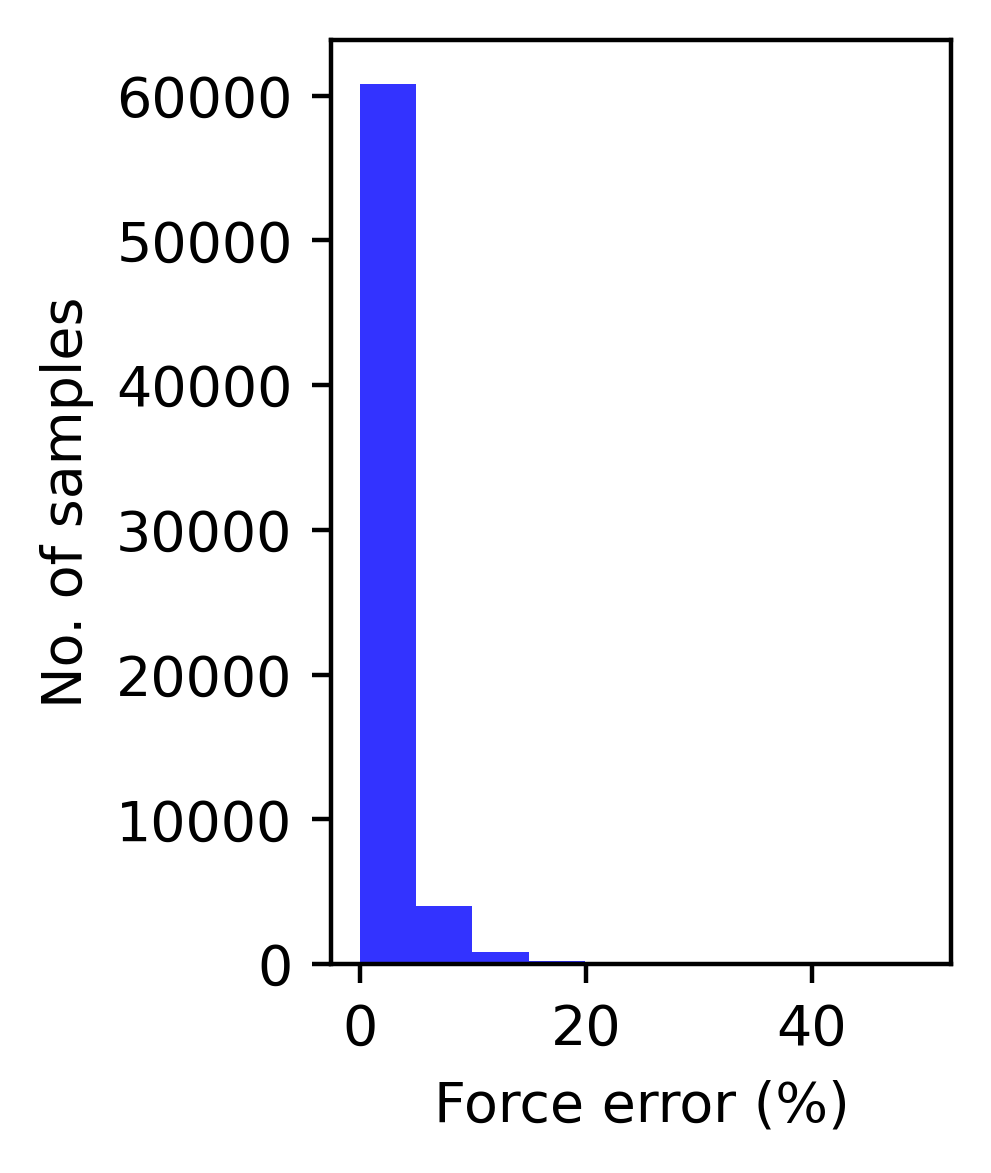}
      \caption{$E_f$ (\%)}
      \label{fig:hist_force_error_scaled_face1}
      \vspace{-2pt}
      \end{subfigure}
      \begin{subfigure}{0.21\linewidth}
      \includegraphics[width=1.0\linewidth]{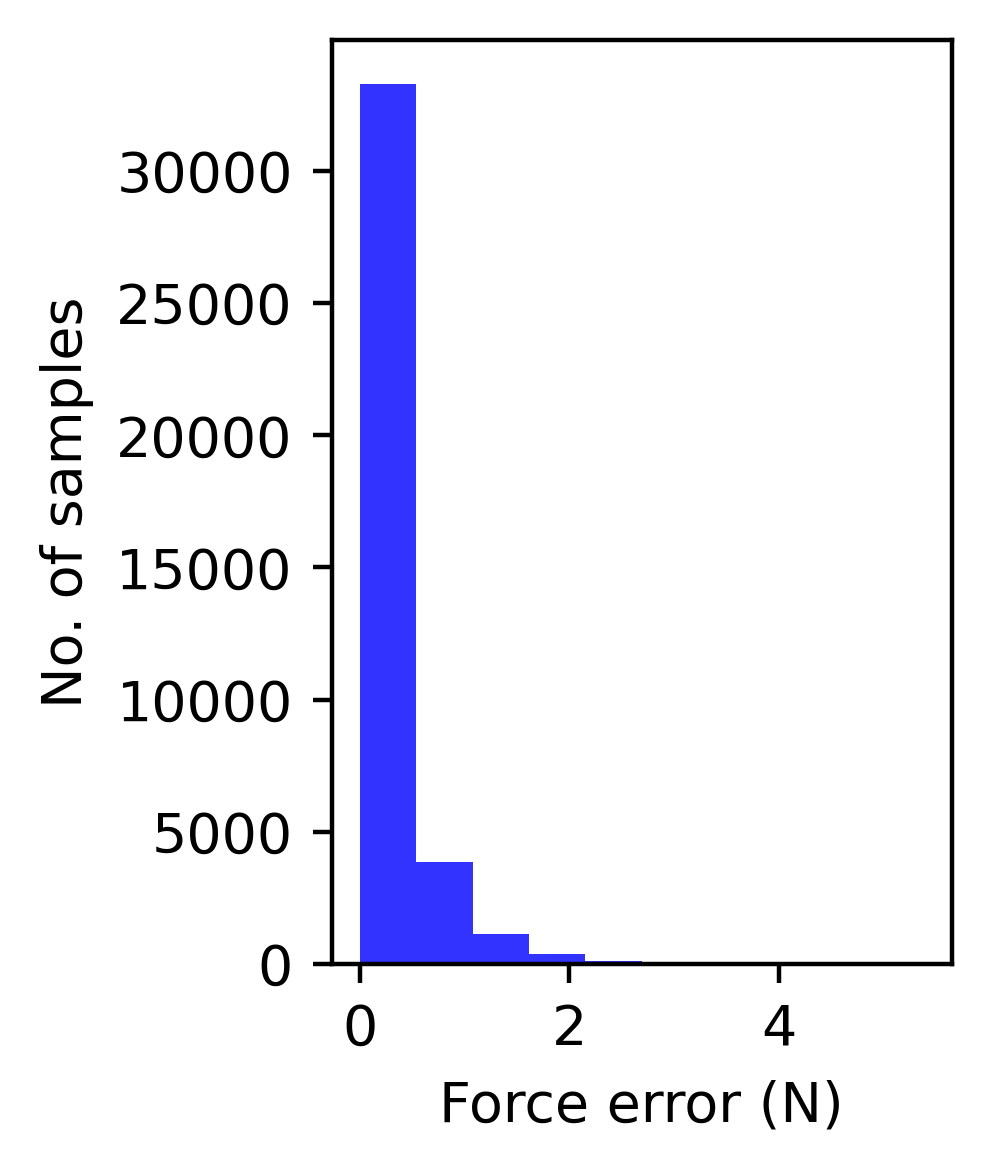}
      \caption{$E_f$ (N)}
      \label{fig:hist_force_error_face1}
      \vspace{-2pt}
      \end{subfigure}
      \begin{subfigure}{0.33\linewidth}
      \includegraphics[width=1.0\linewidth]{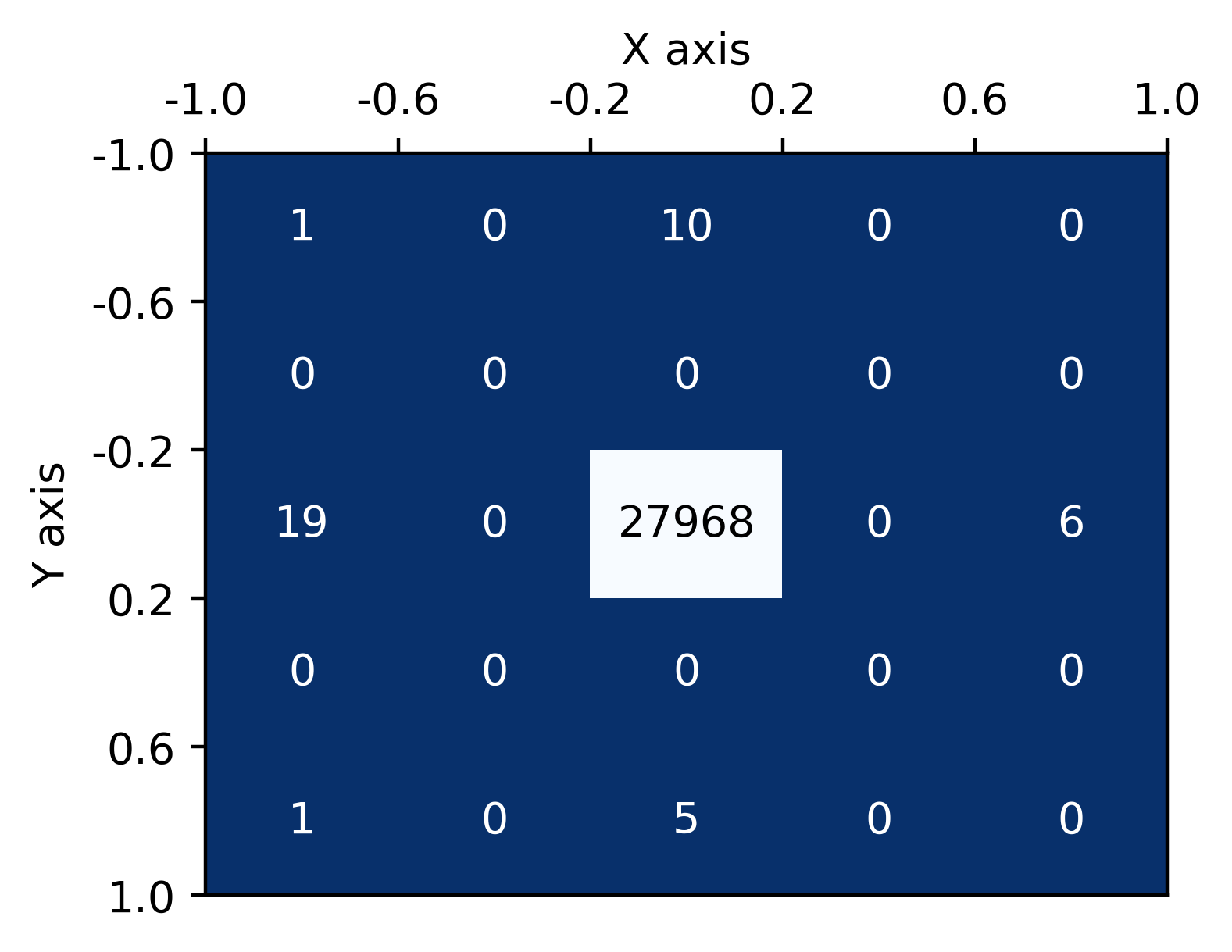}
      \caption{$E_{loc}$}
      \label{fig:hist_loc_error_face1}
      \vspace{-2pt}
      \end{subfigure}
      \caption{\textbf{Face1} histograms of $A_{sim}$, $E_f$ and $E_{loc}$. Two histograms are shown for $E_f$ in both percentage (b) and real (c) errors. 
      (d) The spatial distribution of $E_{loc}$.
      }
      \label{fig:hist_metrics_face1}
      \vspace{-10pt}
\end{figure}

\begin{figure*}[thpb]
      \centering
      \vspace{2pt}
      \begin{subfigure}{0.3135\linewidth}
      \includegraphics[width=1.0\linewidth]{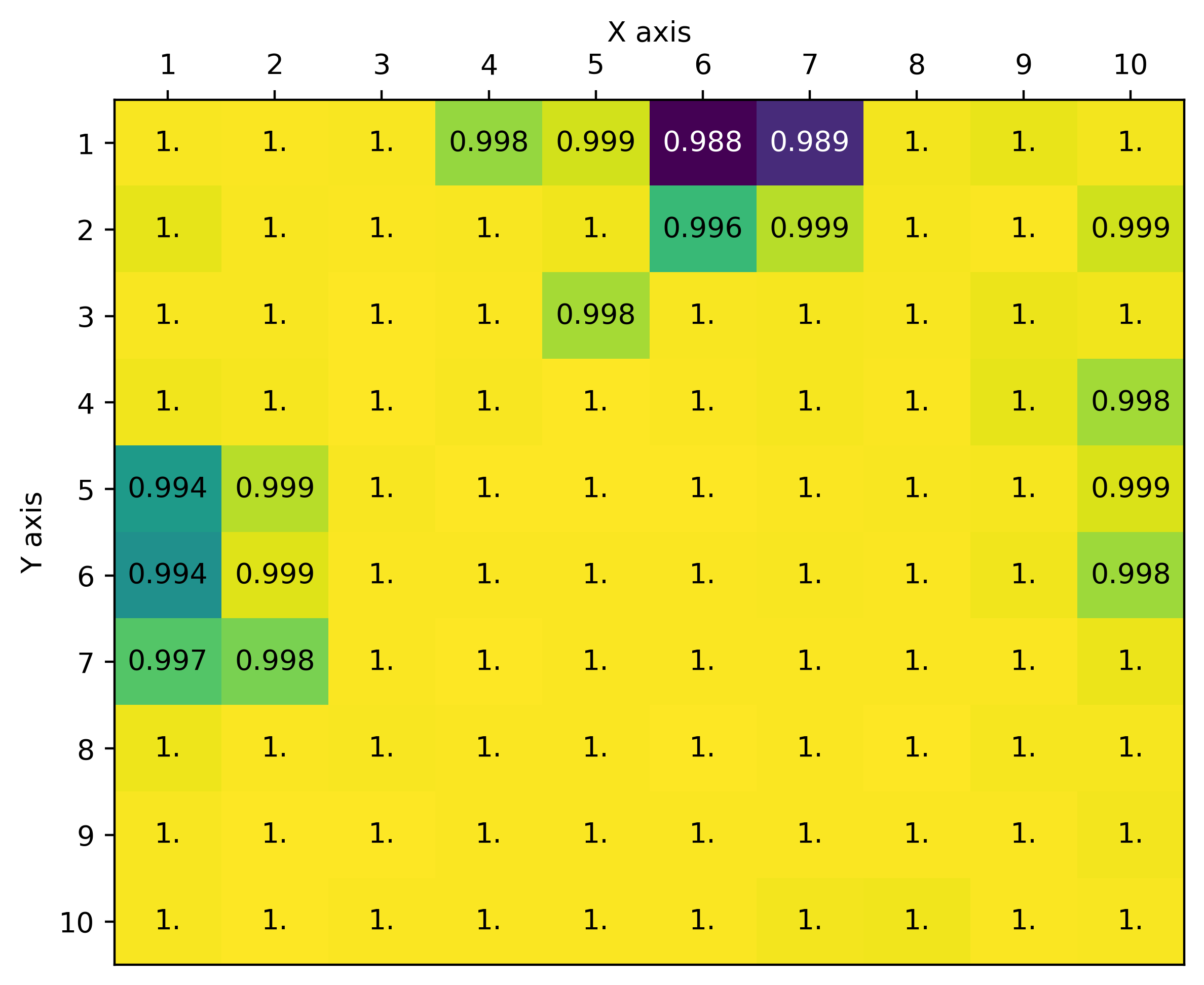}
      \caption{Average $A_{sim}$}
      \label{fig:loc_wise_sim}
      \vspace{-2pt}
      \end{subfigure}
      \begin{subfigure}{0.3135\linewidth}
      \includegraphics[width=1.0\linewidth]{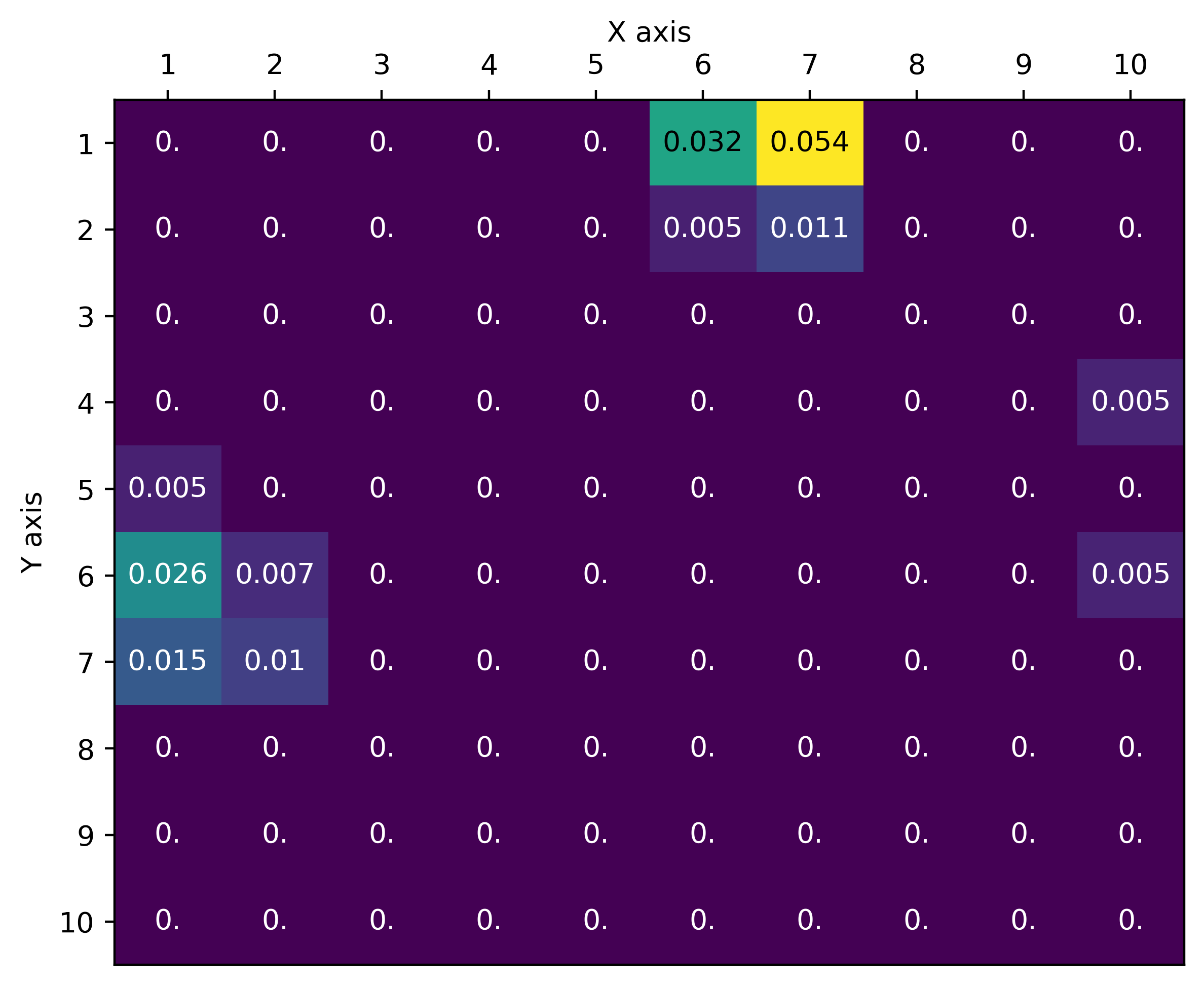}
      \caption{Average $E^{Eucl}_{loc}$}
      \label{fig:loc_wise_loc_error}
      \vspace{-2pt}
      \end{subfigure}
      \begin{subfigure}{0.36\linewidth}
      \includegraphics[width=1.0\linewidth]{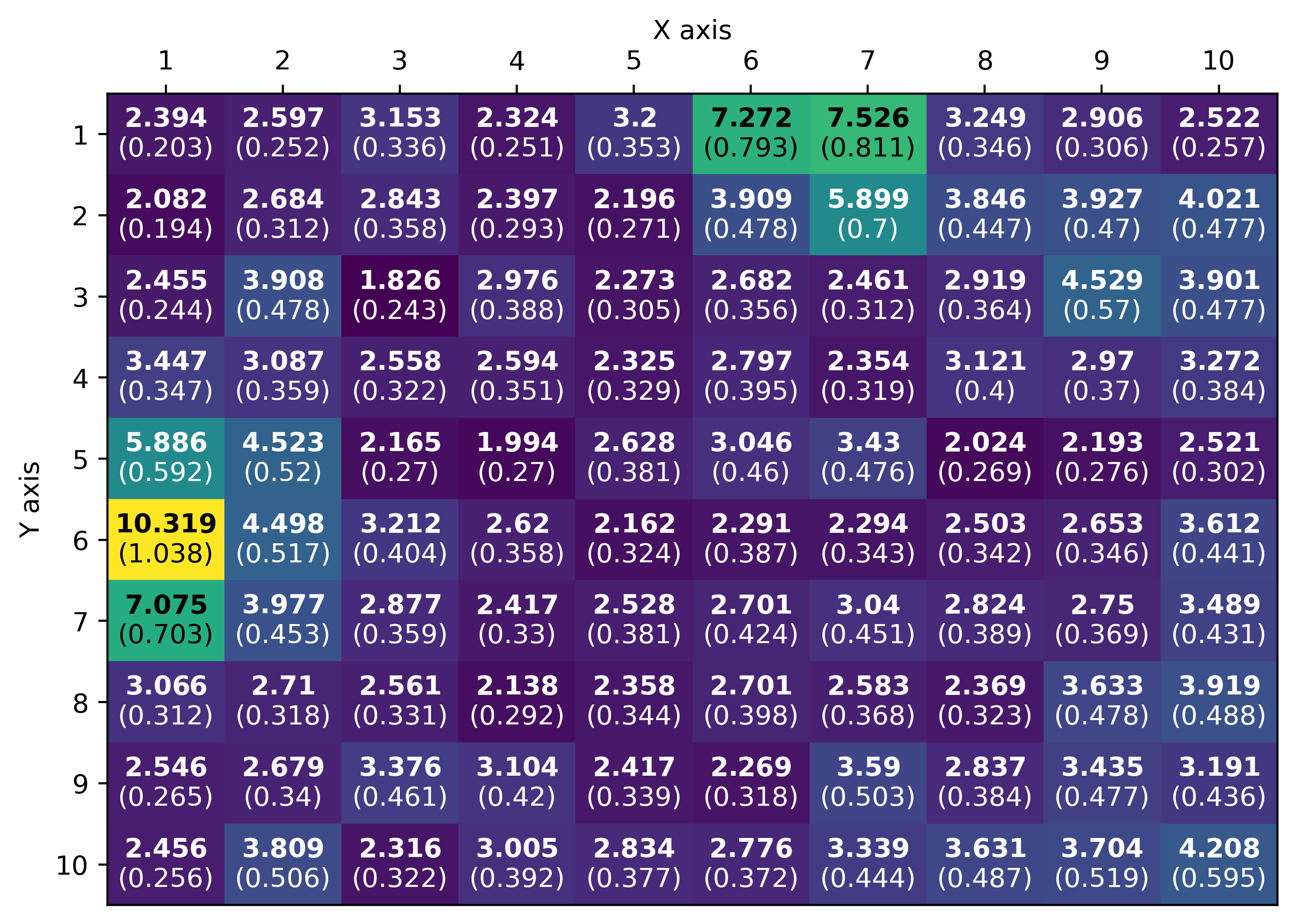}
      \caption{Average $E_{f}$}
      \label{fig:loc_wise_force_error}
      \vspace{-2pt}
      \end{subfigure}
      \caption{\textbf{Face1} location-wise performance. Both percentage (\textbf{bold}) and real force errors (in brackets) are shown in (c).}
      \label{fig:loc_wise_performance}
      \vspace{-10pt}
\end{figure*}

\subsection{Metrics}
The performance of the design was evaluated by measuring the accuracy of estimated contact locations and force magnitudes and also its capability of estimating non-contact cases, using the metrics below.
\par \textit{a) Heatmap similarity} $A_{sim}$: The similarity between the predicted and ground-truth heatmaps is measured via the maximum Zero Normalized Cross Correlation (ZNCC)~\cite{duda1973pattern}.
\par \textit{b) Contact location error} $E_{loc}$: The contact location error is the displacement of the maximum ZNCC between the predicted and ground-truth heatmaps. Both the errors in X-axis ($E^{X}_{loc}$), Y-axis ($E^{Y}_{loc}$), and Euclidean distance ($E^{Eucl}_{loc}$) are calculated.
\par \textit{c) Force magnitude error} $E_{f}$: The magnitude error is the difference between predicted and ground-truth force of contact locations (positive in ground-truth). Both full-scale percentage (divided by the force ranges in Fig.~\ref{fig:loc_wise_force_range}) and errors in Newtons are measured.
\par \textit{d) Non-contact accuracy} $A_{non}$: The non-contact accuracy measures the binary prediction accuracy of non-contact cases (no contact point across all locations in ground-truth). A heatmap prediction is classified as non-contact if all its elements have values smaller than a threshold. The threshold is defined in this work as 90\% of the minimum contact force across the whole dataset $\mathbf{D}_I$.

\subsection{Contact estimation performance}
\label{sec:general_performance}

\begin{table}[]
    \centering
    \setlength{\tabcolsep}{4.5pt}
    \begin{tabular}{c|c|c|c|c}
    \hline
         Face & Avg. $A_{sim}$ & Avg. $E_{loc}$ ($\times 10^{-4}$) & Avg. $E_{f}$ & $A_{non}$\\
    \hline
         \textbf{Face1} & 0.999 & 15.3 (9.64, 6.07)& 1.82\% (0.278N) & 1.0\\
    \hline
         \textbf{Face2} & 0.999 & 67.8 (2.50, 65.3)& 2.09\% (0.380N) & 1.0\\
    \hline
         \textbf{Face3} & 0.997 & 127 (35.7, 96.0)& 3.23\% (0.479N) & 1.0\\
    \hline
         \textbf{Face4} & 0.999 & 60.4 (15.0, 47.1)& 1.99\% (0.370N) & 1.0\\
    \hline
         \textbf{Face5} & 0.999 & 25.5 (10.0, 15.7)& 2.09\% (0.410N) & 1.0\\
    \hline
    \end{tabular}
    \vspace{-2pt}
    \caption{Average performance of different faces with single, dual, and triple contact points. The average Euclidean location errors $E^{Eucl}_{loc}$ are listed in the unit of $\times 10^{-4}$pixels, with their errors in X- and Y-axis in brackets ($E^{X}_{loc}$, $E^{Y}_{loc}$). The average force errors $E_{f}$ are listed in percentage with their values in Newtons in brackets.}
    \label{tab:general_performance}
    \vspace{-10pt}
\end{table}

This section evaluates the performance of the design with the models trained with the settings in Section~\ref{sec:training}. Table~\ref{tab:general_performance} shows the average performance in the three metrics of the five faces. We can see that the contact estimation was remarkably accurate from all metric perspectives, which is also shown by the histograms of \textbf{Face1} in Fig.~\ref{fig:hist_metrics_face1}. The average force errors were within 3.23\% and 0.479N (very small when considering the ground-truth force resolution was $\pm$0.12N), indicating that the design can accurately estimate single and multiple contact points on one face for all five faces.

\subsection{Location-wise performance}
\label{sec:loc_wise_eval}

\begin{figure}
    \centering
    \includegraphics[width=0.85\linewidth]{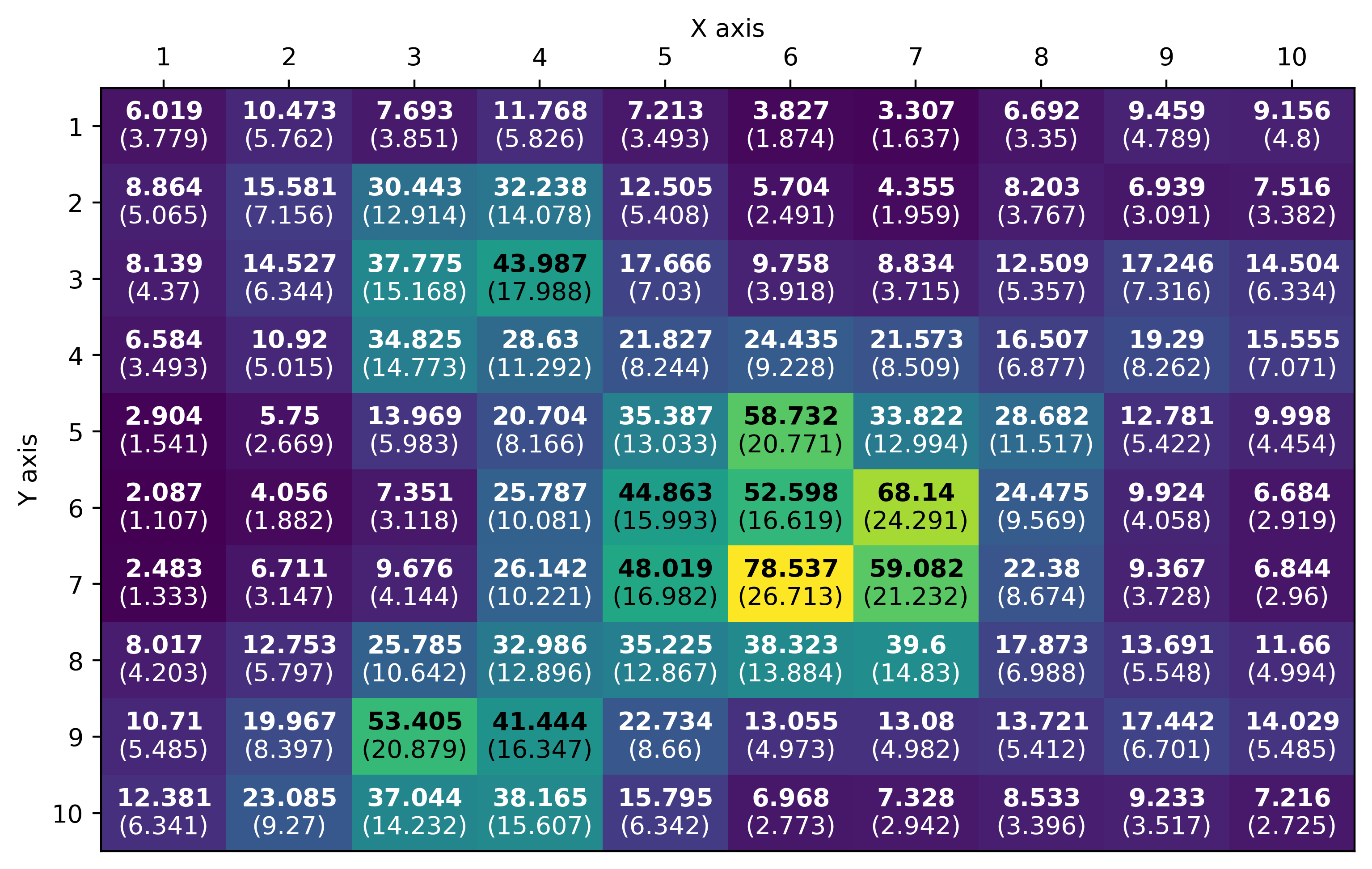}
    \caption{\textbf{Face1} location-wise force sensitivity for single-contact cases, with \textbf{bold} numbers in the normalized scale ($\times 10^{-15}$) and numbers in brackets for the real scale ($\times 10^{15}$).}
    \label{fig:loc_wise_hall_vol_face1}
    \vspace{-8pt}
\end{figure}

\begin{figure*}[thpb]
      \centering
      \vspace{2pt}
      \begin{subfigure}{0.3135\linewidth}
      \includegraphics[width=1.0\linewidth]{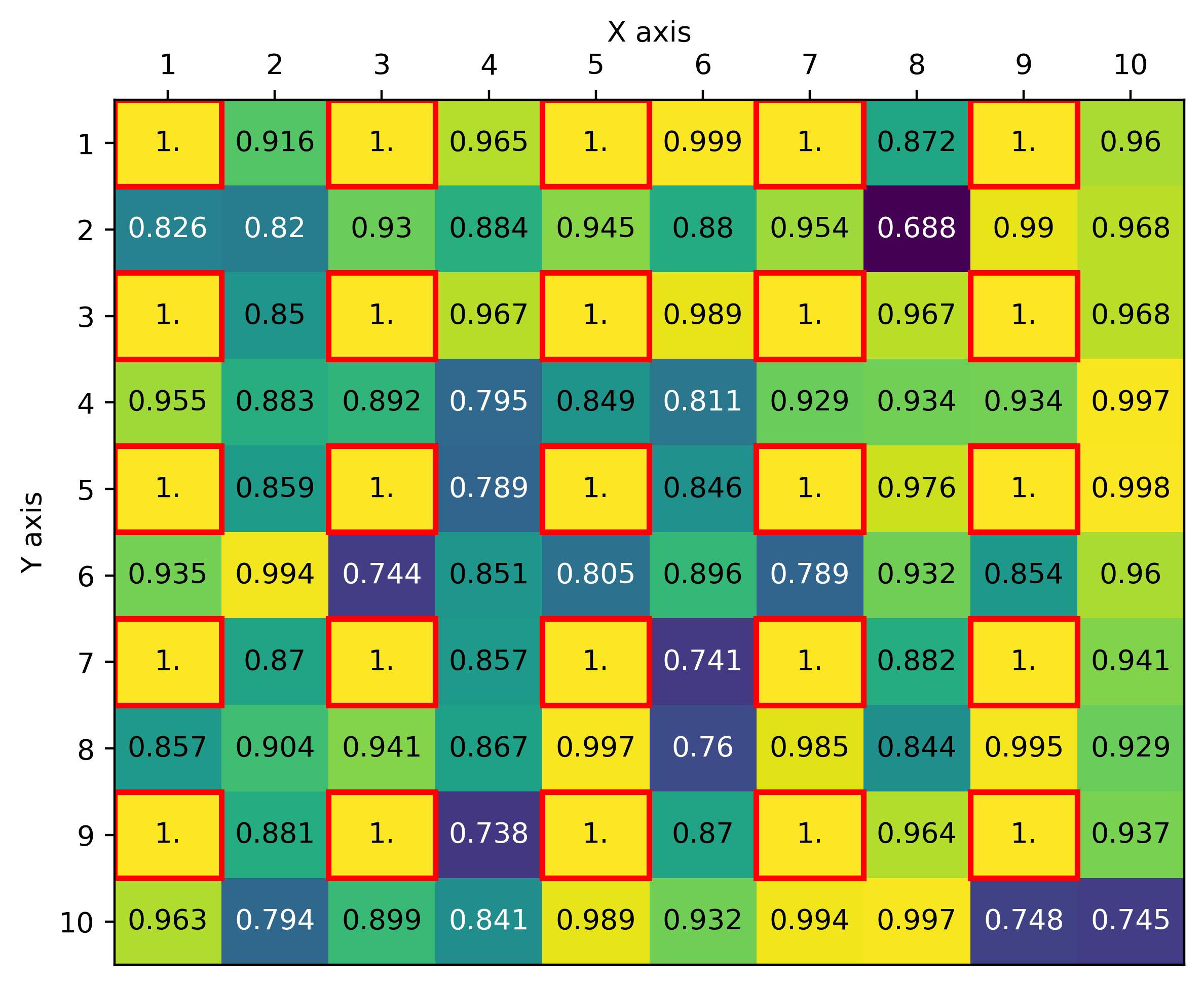}
      \caption{Average $A_{sim}$}
      \label{fig:seen_vs_unseen_sim}
      \vspace{-2pt}
      \end{subfigure}
      \begin{subfigure}{0.3135\linewidth}
      \includegraphics[width=1.0\linewidth]{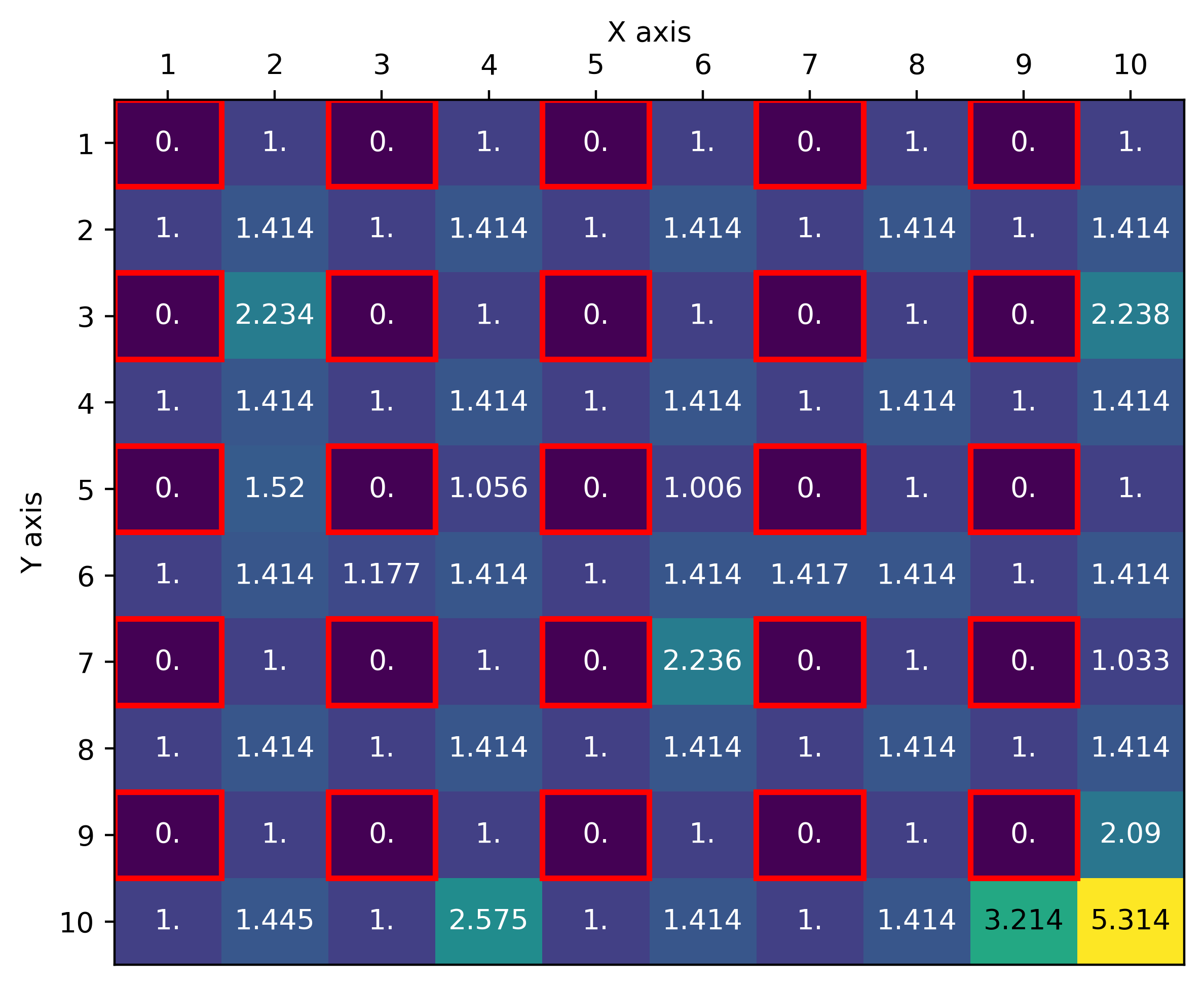}
      \caption{Average $E^{Eucl}_{loc}$}
      \label{fig:seen_vs_unseen_loc_error}
      \vspace{-2pt}
      \end{subfigure}
      \begin{subfigure}{0.36\linewidth}
      \includegraphics[width=1.0\linewidth]{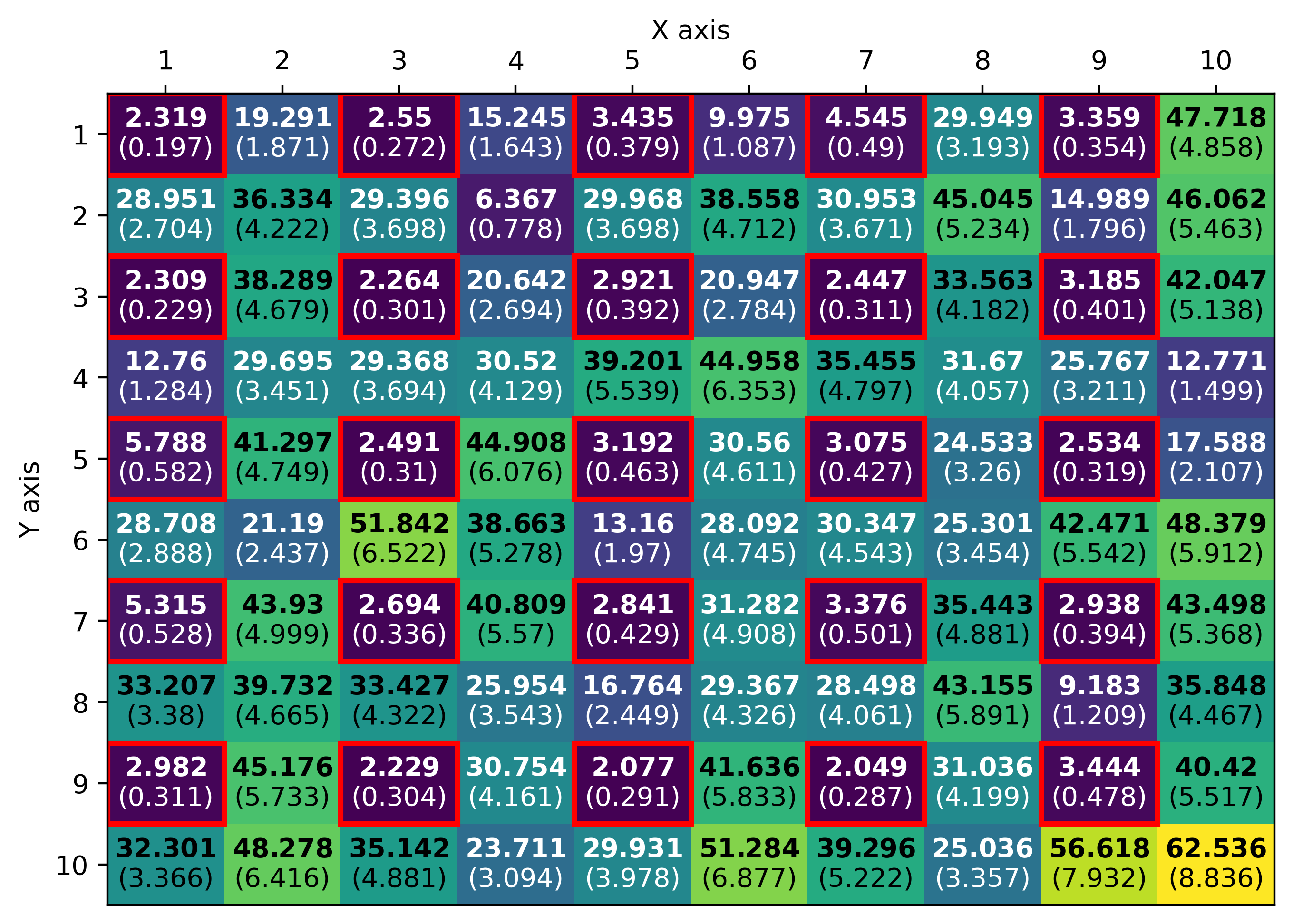}
      \caption{Average $E_{f}$}
      \label{fig:seen_vs_unseen_force_error}
      \vspace{-2pt}
      \end{subfigure}
      \caption{Performance in seen (marked with red rectangles) and unseen locations. 
      Performance degradation in unseen locations was significant in $E_{f}$ (mostly $>$10\%), but relatively small in $E^{Eucl}_{loc}$ (mostly $\leqslant$1.414 pixels) and $A_{sim}$ (mostly $>$0.8).}
      \label{fig:seen_vs_unseen}
      \vspace{-5pt}
  \end{figure*}

  \begin{figure*}
        \vspace{-3pt}
      \centering
      \begin{subfigure}{0.329\linewidth}
      \includegraphics[width=1.0\linewidth]{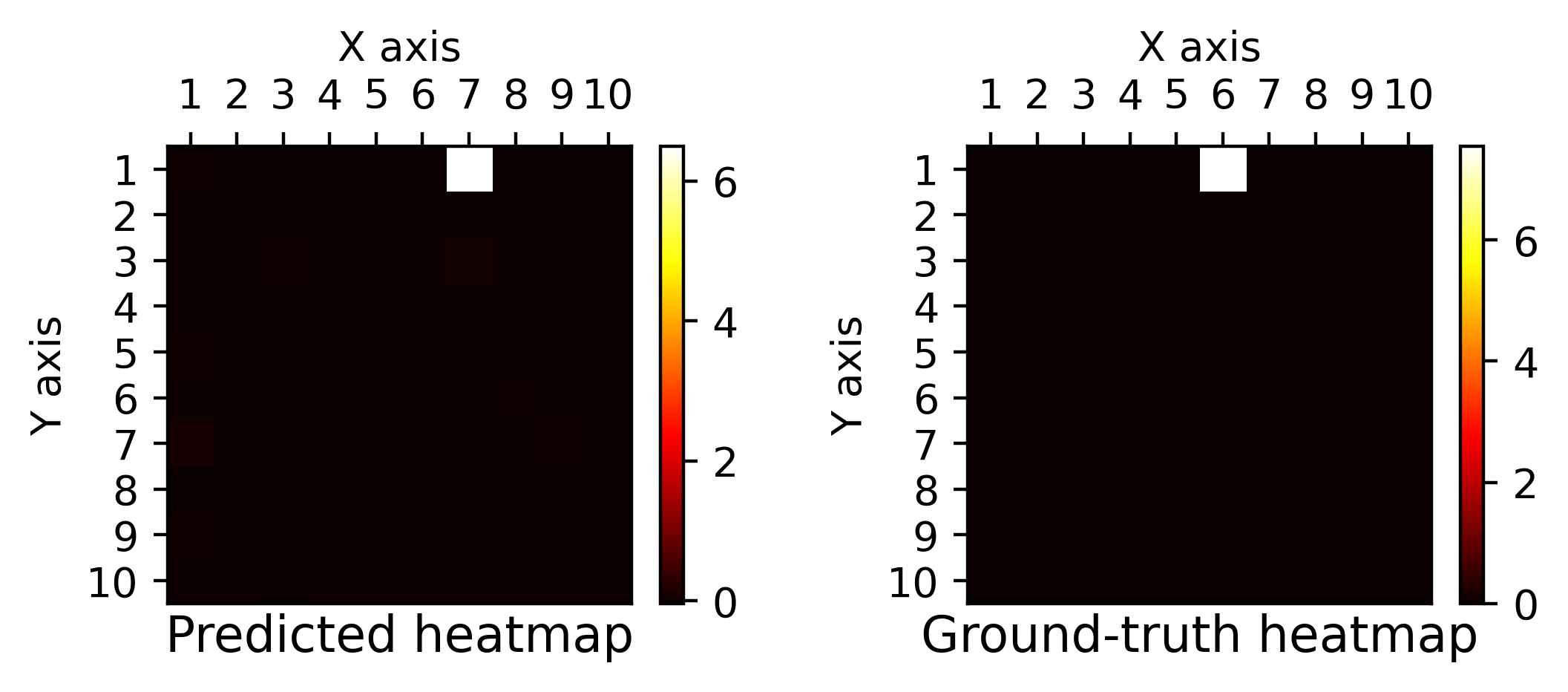}
      \caption{Normal case 1}
      \label{fig:normal_case1}
      \vspace{-2pt}
      \end{subfigure}
      \begin{subfigure}{0.329\linewidth}
      \includegraphics[width=1.0\linewidth]{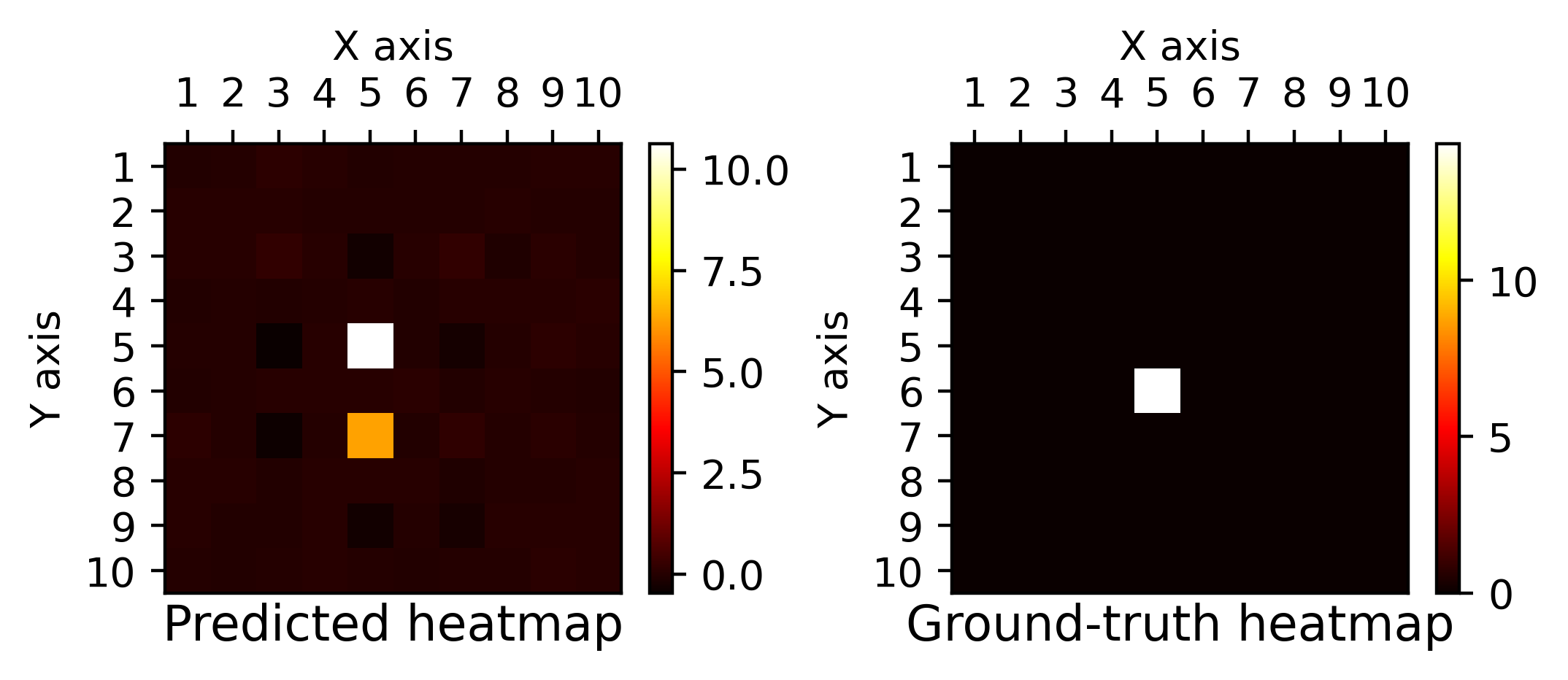}
      \caption{Normal case 2}
      \label{fig:normal_case2}
      \vspace{-2pt}
      \end{subfigure}
      \begin{subfigure}{0.329\linewidth}
        \includegraphics[width=1.0\linewidth]{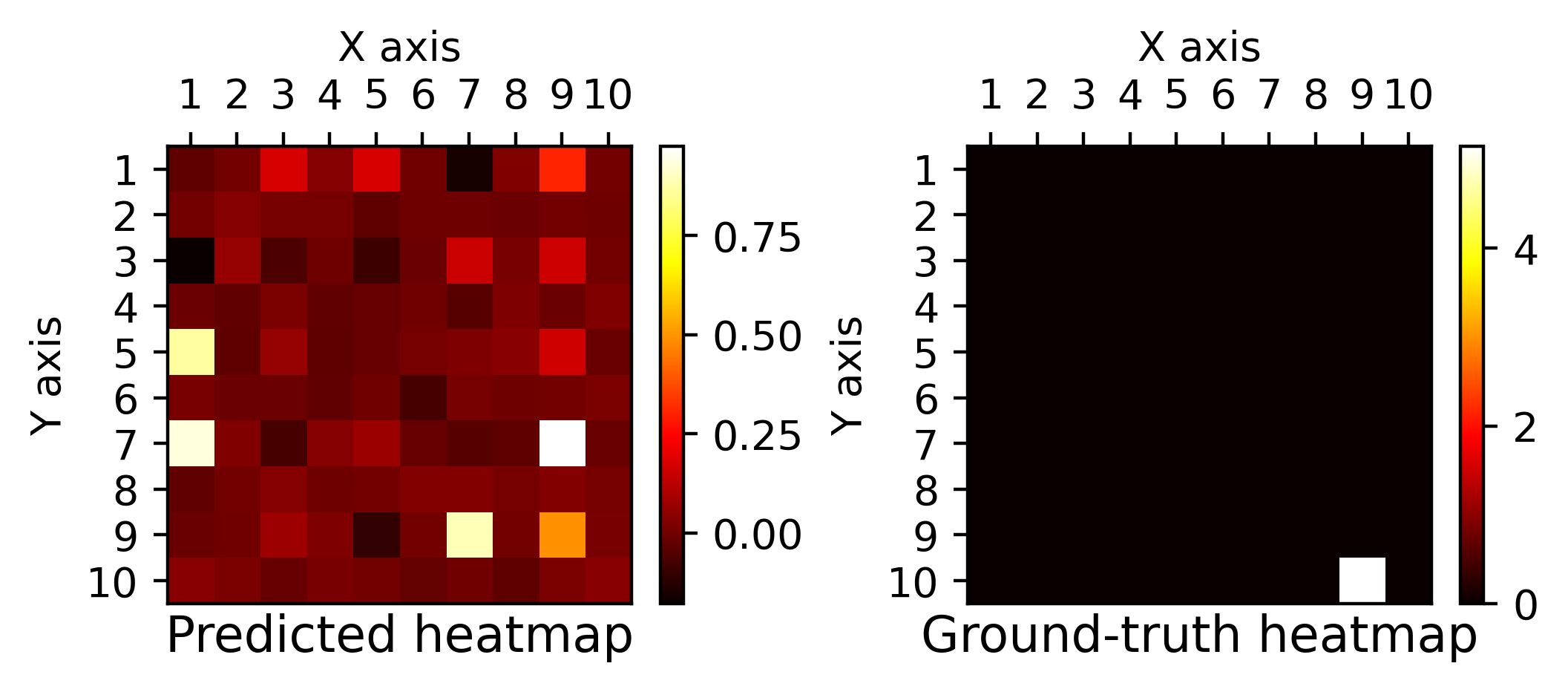}
      \caption{Outlier case}
      \label{fig:outlier_case}
      \vspace{-2pt}
      \end{subfigure}
      \caption{Typical heatmaps for unseen cases. In most cases, nearby seen locations were estimated for unseen locations (a and b), but predictions could be noisier in some outlier cases (c). More results are shown in the video: \href{https://youtu.be/kQSZlNxYRrs}{https://youtu.be/kQSZlNxYRrs}.}
      \label{fig:typical_samples_unseen}
      \vspace{-12pt}
  \end{figure*}

  This section evaluates how the design works at each pixel location with the same models in Section~\ref{sec:general_performance}. Fig.~\ref{fig:loc_wise_performance} shows the average performance for all locations of \textbf{Face1} in single-contact situations (dual or triple contact situations are not included as they could bring in unnecessary biases as not all locations were covered by multi-contact situations as shown in Fig.~\ref{fig:data_coverage}). From the figure we can see that errors ($E^{Eucl}_{loc}$ and $E_f$) or small similarities ($A_{sim}$) appeared in locations in the mid-edge areas, which is not surprising as the Hall signal changes caused by contacts in those areas could be relatively smaller than the other areas. This is determined by the locations of the magnets and Hall sensors in the design (shown in Fig.~\ref{fig:InternalShell}). The results for the other four faces are similar to \textbf{Face1} presented here, please refer to the project page for the results of other faces.

  To better understand the relation between the location-wise performance and the design, we visualize in Fig.~\ref{fig:loc_wise_hall_vol_face1} the location-wise force sensitivity $\delta$ w.r.t. Hall signal convex hull volume: $\delta = \frac{ConvexHullVolume(\mathbf{S})}{range(\mathbf{F})}$, where $range(\cdot)$ returns the magnitude range of the applied force $\mathbf{F} \in \mathbb{R}^{N}$, and $ConvexHullVolume(\cdot)$ calculates the volume of the convex hull~\cite{de2000computational} for the Hall signals $\mathbf{S} \in \mathbb{R}^{N\times9}$ ($N$ is the number of frames). 
  The force sensitivity indicates how many Hall signal changes are caused by the same amount of force. 
  From Fig.~\ref{fig:loc_wise_hall_vol_face1}, we can observe the high correlation between the force sensitivity and all three metrics, which further shows that the locations of magnets and sensors determine how Hall signals respond to the contacts from different locations, and therefore influencing the contact estimation performance: the higher the force sensitivity is, the better the performance will be. Future designs need to consider more the magnetic field distributions w.r.t. Hall sensors for a more uniform force sensitivity.

\subsection{Performance in unseen contact locations}
\label{sec:unseen_vs_seen}

\begin{table}[]
    \centering
    \begin{tabular}{c|c|c|c}
    \hline
         Scenario & Avg. $A_{sim}$ & Avg. $E_{loc}$ (pixel) & Avg. $E_{f}$\\
    \hline
         \textbf{Seen} & 1.00 & 0 (0, 0)& 3.06\% (0.372N) \\
    \hline
         \textbf{Unseen} & 0.896 & 1.32 (0.706, 0.858)& 32.6\% (4.18N) \\
    \hline
    \end{tabular}
    \vspace{-2pt}
    \caption{Average performance in seen (marked with red rectangles in Fig.~\ref{fig:seen_vs_unseen}) and unseen locations on \textbf{Face1}.
    }
    \label{tab:seen_vs_unseen}
    \vspace{-10pt}
\end{table}

To study how the design would work in unseen contact locations, an experiment was conducted to train a model for \textbf{Face1} with only the data from a subset of the locations, assuming the instrumented object was calibrated using data collected with a lower resolution (from $10\times10$ to $5\times5$). Specifically, the model was trained with many fewer contact locations: only the ones with red rectangles in Fig.~\ref{fig:seen_vs_unseen}, the other locations were unseen in training. 

In the experiment, only single-contact $\mathbf{D}^{1}_{1}$ (100k samples: 25k seen $\mathbf{D}^{1}_{1,seen}$, 75k unseen $\mathbf{D}^{1}_{1,unseen}$) and $\frac{1}{3}$ non-contact ${\mathbf{D}^{0}_{1}}^{\frac{1}{3}}$ (1k samples) data was used. Multiple-contact data was not used to avoid unnecessary biases, as not all locations were covered by multiple-contact cases. 
A model was trained using 60\% of $\mathbf{D}^{1}_{1,seen}$ (15k samples) and 60\% of ${\mathbf{D}^{0}_{1}}^{\frac{1}{3}}$ (0.6k samples). The validation set consists of 20\% of $\mathbf{D}^{1}_{1,seen}$ and 20\% of ${\mathbf{D}^{0}_{1}}^{\frac{1}{3}}$ (5.2k samples in total). The testing set is composed of two parts: the remaining 20\% of $\mathbf{D}^{1}_{1,seen}$ and ${\mathbf{D}^{0}_{1}}^{\frac{1}{3}}$, and all of $\mathbf{D}^{1}_{1,unseen}$ (80.2k samples in total). Other settings are the same as in Section~\ref{sec:training}.

Table~\ref{tab:seen_vs_unseen} lists the average $A_{sim}$, $E_{loc}$ and $E_{f}$. Fig.~\ref{fig:seen_vs_unseen} shows its location-wise performance. From both the overall and location-wise results, we can see that the model performs well in seen locations (similar to that in Section~\ref{sec:general_performance}). However, its performance in unseen locations is significantly worse in force magnitude errors (increased from 3.06\% to 32.6\%), while the performance change in $A_{sim}$ and $E_{loc}$ is relatively small. Even in unseen locations, the model is able to output heatmaps very similar to ground-truth ones (with average similarities larger than 0.8 in 65 out of 75 locations). It also has average contact location errors smaller than 1.414 pixels in 65 locations. In most cases, the predicted contact locations for unseen cases were nearby seen locations (Fig.~\ref{fig:normal_case1} and~\ref{fig:normal_case2}, while there were some outlier cases (Fig.~\ref{fig:outlier_case}). This indicates that the coverage of training data on more locations is crucial for better contact estimation accuracy, particularly for smaller force magnitude errors.

\subsection{Amount of training data vs performance}
\label{sec:data_amount}
\begin{figure}
    \centering
    \begin{subfigure}{0.488\linewidth}
      \includegraphics[width=1.0\linewidth]{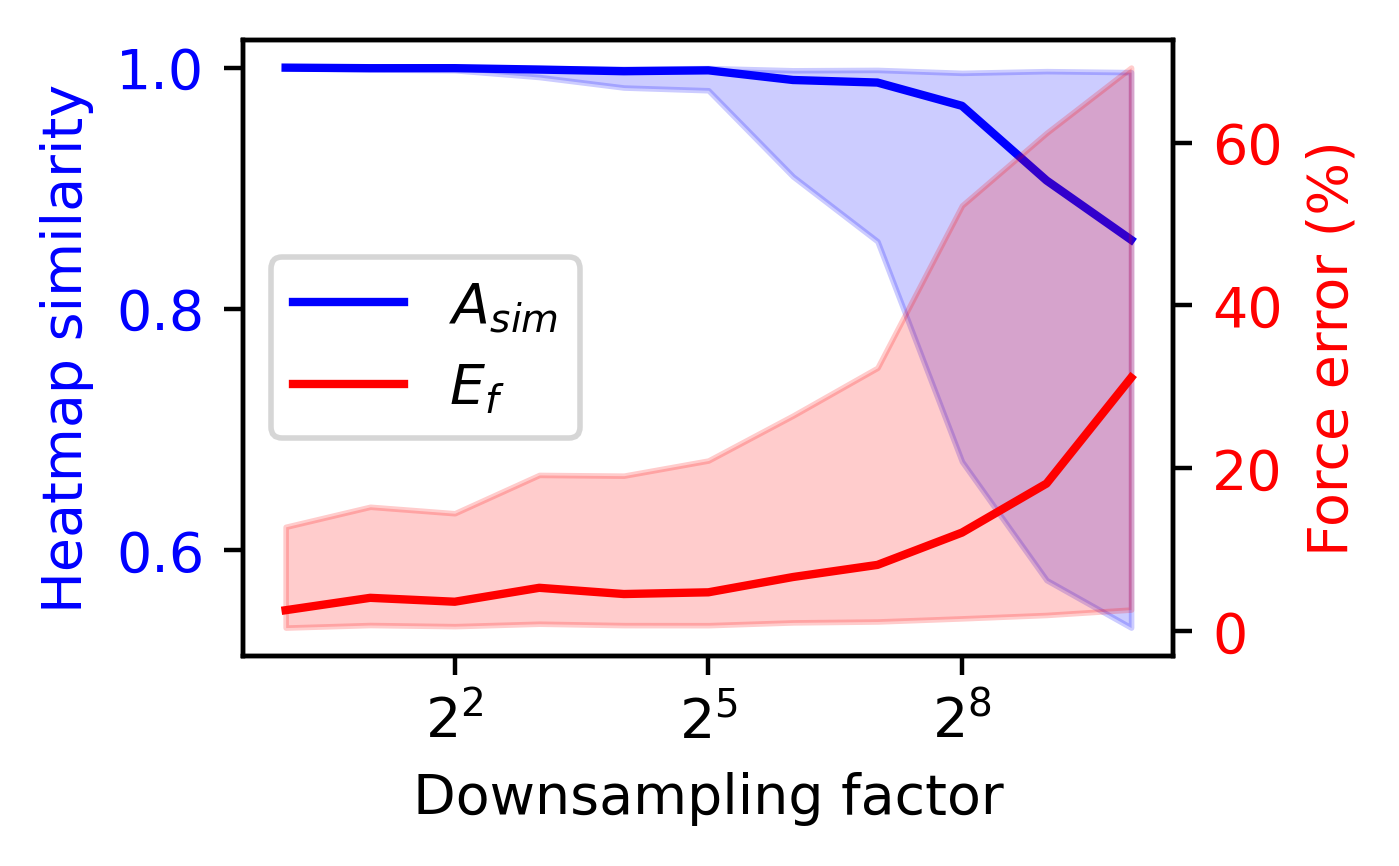}
      \caption{$A_{sim}$ and $E_{f}$}
      \label{fig:data_ablation_sim}
      \vspace{-2pt}
      \end{subfigure}
      \begin{subfigure}{0.499\linewidth}
      \includegraphics[width=1.0\linewidth]{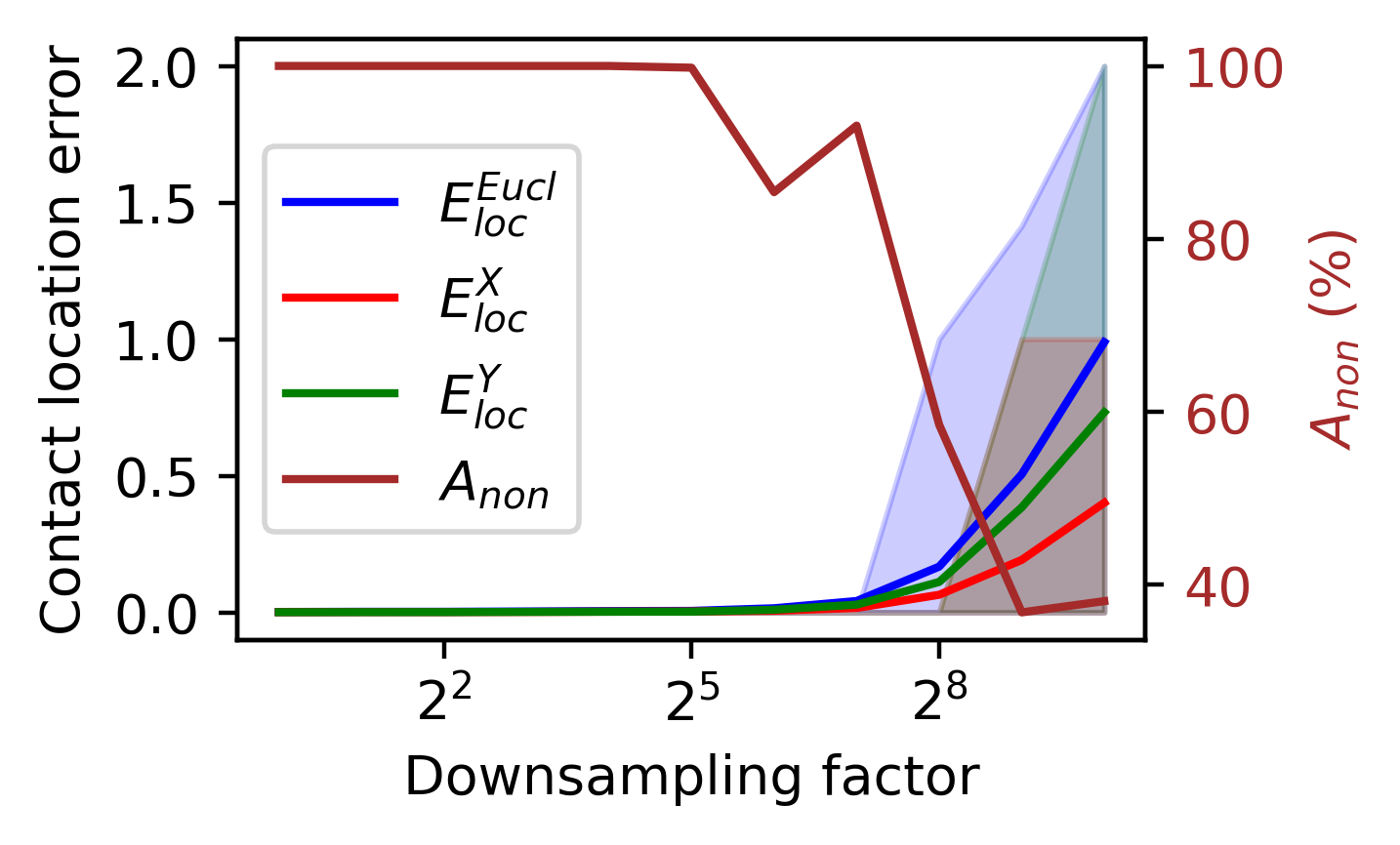}
      \caption{$E_{loc}$ and $A_{non}$}
      \label{fig:data_ablation_loc}
      \vspace{-2pt}
      \end{subfigure}
    \caption{Performance changes in $A_{sim}$, $E_f$, $E_{loc}$, and $A_{non}$ w.r.t. the amount of training data (downsampled by the factors $2^k$, where $k \in [0, 1, ..., 10]$). 
    The curves are shown with their distributions (median, and the 10th and 90th percentiles), except the scalar $A_{non}$. The medians (mostly zeros) of $E_{loc}$ errors are replaced with their average values in (b) to better show the performance changes.
    }
    \label{fig:training_data_ablation}
    \vspace{-10pt}
\end{figure}

This section explores how the amount of training data would influence the contact estimation performance. In the experiment, ten more models were trained for \textbf{Face1} in addition to the one in Section~\ref{sec:general_performance}, but with fewer samples: the original training set $\mathbf{D}_{1,\mbox{Tr}}$ was downsampled by factors of $2^k, k=[1, 2, \ldots, 10]$. The other settings are the same as in Section~\ref{sec:training}. Fig.~\ref{fig:training_data_ablation} shows the performance changes w.r.t. the amount of training data (indicated by downsampling factors). We can see the performance had obvious degradation after the downsampling factor $2^5$ (2683 training samples in total, and about 19 samples for each case) for all metrics ($A_{sim}$, $E_f$, $E_{loc}$, and $A_{non}$). This indicates that as few as 2683 samples can enable accurate contact estimation.

\section{Evaluation in Real Applications}
\label{sec:eval_real}

\begin{figure}
    \centering
    \begin{subfigure}{0.462\linewidth}
      \includegraphics[width=1.0\linewidth]{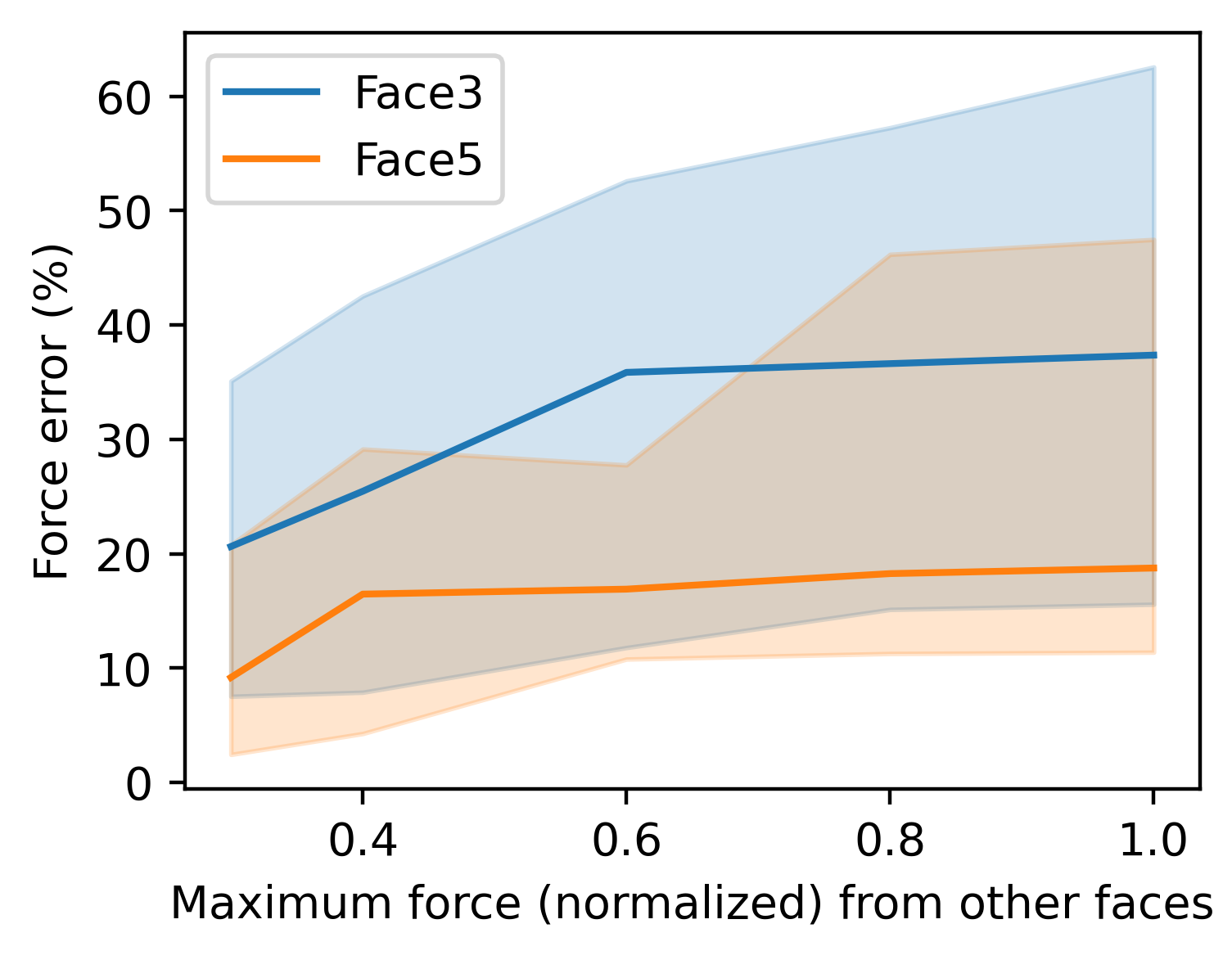}
      \caption{Contact faces}
      \label{fig:perf_vs_force_range_contact}
      \vspace{-2pt}
      \end{subfigure}
      \begin{subfigure}{0.5252\linewidth}
      \includegraphics[width=1.0\linewidth]{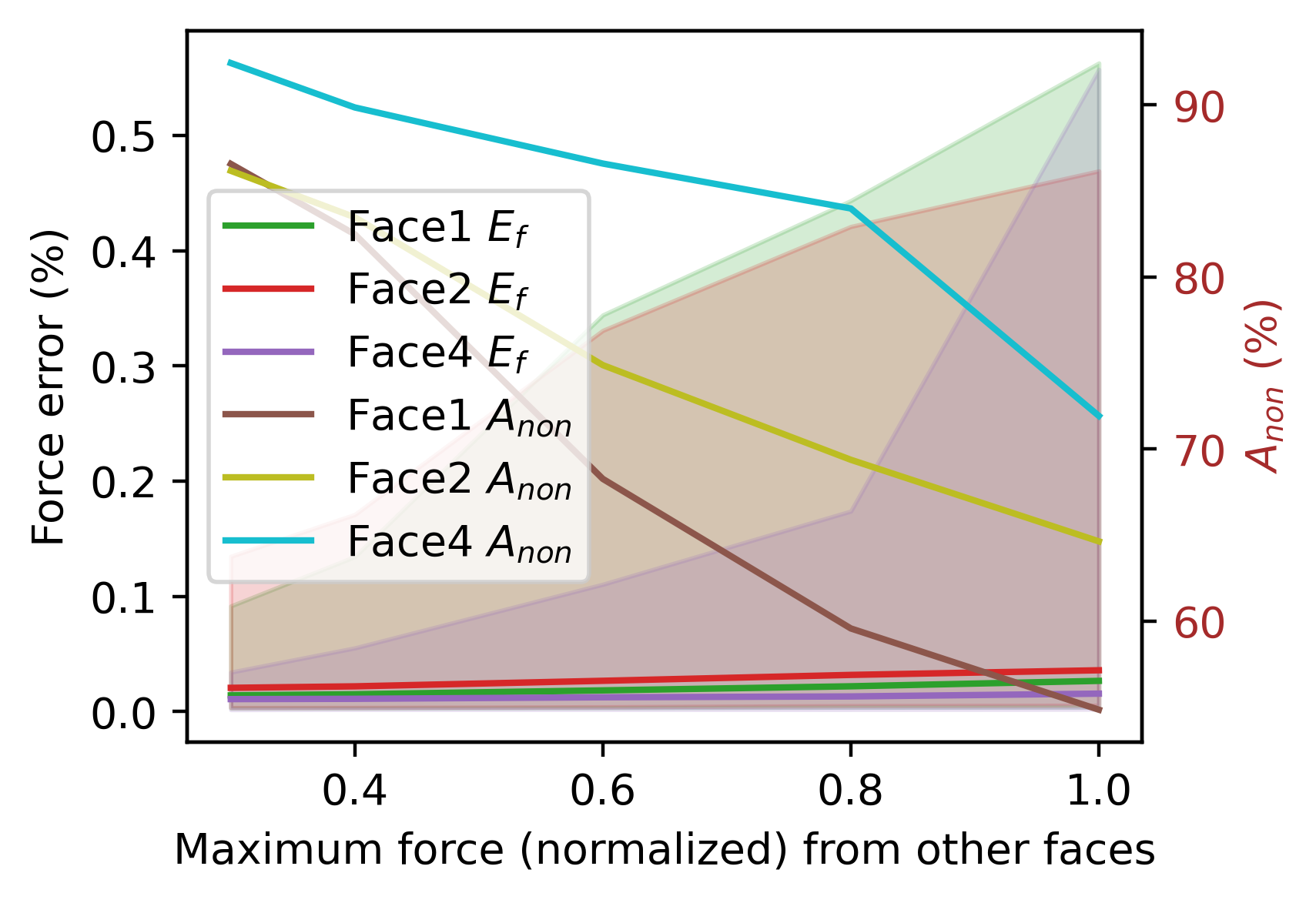}
      \caption{Non-contact faces}
      \label{fig:perf_vs_force_range_noncontact}
      \vspace{-2pt}
      \end{subfigure}
    \caption{Performance changes in $E_f$ and $A_{non}$ w.r.t. the allowed maximum force (ranging from 0.3 to 1.0) from other faces (the faces other than the one being evaluated). The $E_f$ curves are shown with their distributions (median, and the 10th and 90th percentiles).
    }
    \label{fig:performance_vs_force_range}
    \vspace{-5pt}
\end{figure}

\begin{table}
    \centering
    \setlength{\tabcolsep}{5pt}
    \begin{tabular}{c|c|c|c|c}
    \hline
         Face & Avg. $A_{sim}$ & Avg. $E_{loc}$ (pixel) & Avg. $E_{f}$ & $A_{non}$\\
    \hline
         \textbf{Face3} & 0.631 & 1.91 (1.01, 1.27)& 38.7\% (5.76N) & 0.933\\
    \hline
         \textbf{Face5} & 0.678 & 2.96 (1.91, 1.87)& 23.8\% (4.61N) & 0.984\\
    \hline
    \end{tabular}
    \vspace{-2pt}
    \caption{Average performance of \textbf{Face3} and \textbf{Face5}. The average force errors $E_{f}$ are listed in percentage with their values in Newtons in brackets. The average Euclidean location errors $E^{Eucl}_{loc}$ are listed in the unit of pixel, with their errors in the X- and Y-axis in brackets ($E^{X}_{loc}$, $E^{Y}_{loc}$).}
    \label{tab:real_test_parallel}
    \vspace{-10pt}
\end{table}

 \begin{figure*}[!t]
      \centering
      \vspace{2pt}
      \begin{subfigure}{0.329\linewidth}
      \includegraphics[width=1.0\linewidth]{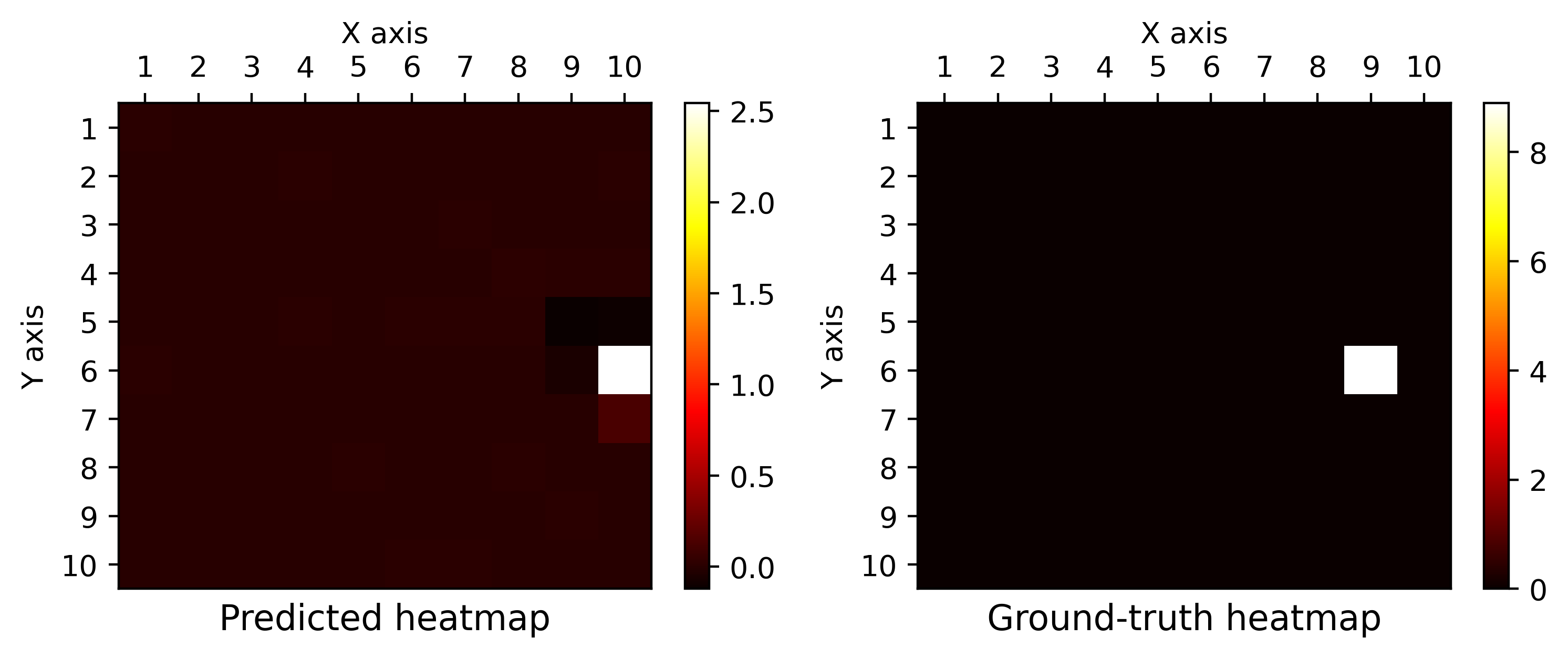}
      \caption{Good case for the single-contact probe}
      \label{fig:good_case1}
      \vspace{-2pt}
      \end{subfigure}
      \begin{subfigure}{0.329\linewidth}
      \includegraphics[width=1.0\linewidth]{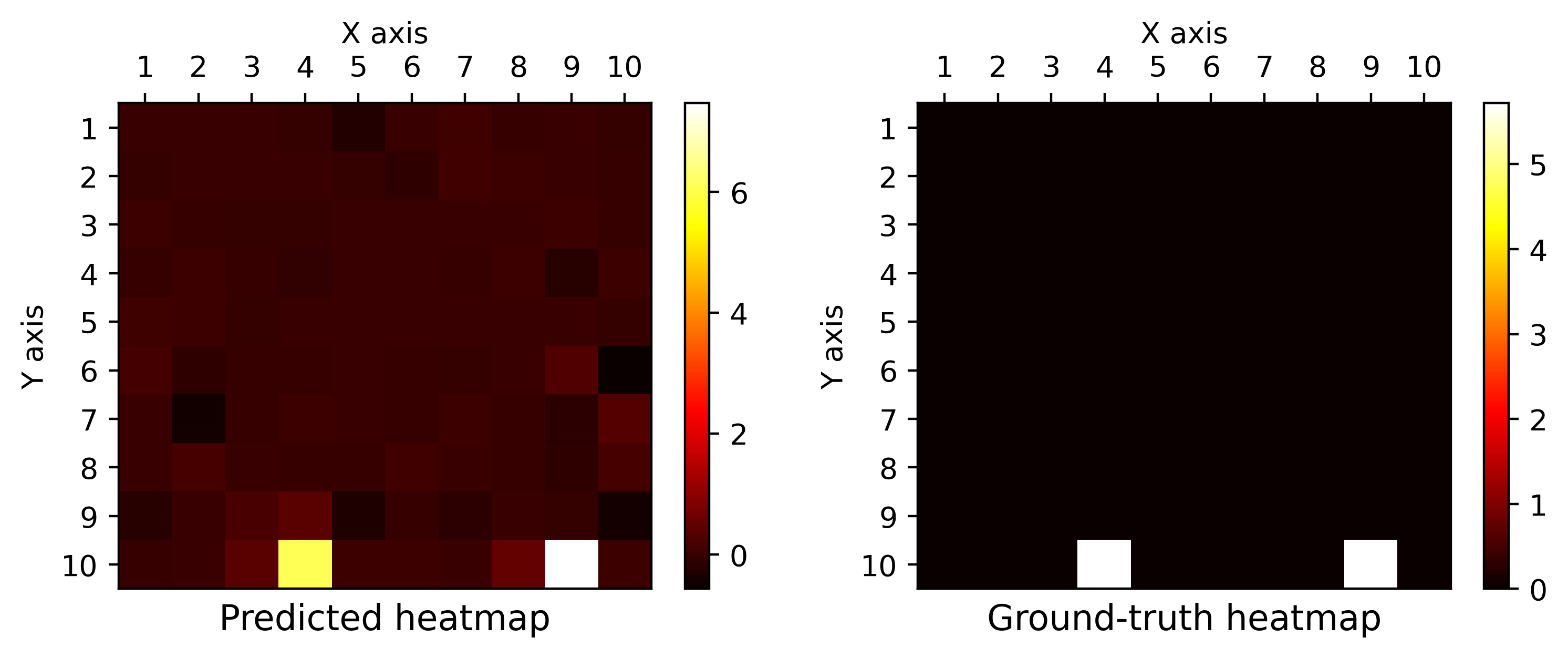}
      \caption{Good case for the dual-contact probe}
      \label{fig:good_case2}
      \vspace{-2pt}
      \end{subfigure}
      \begin{subfigure}{0.329\linewidth}
        \includegraphics[width=1.0\linewidth]{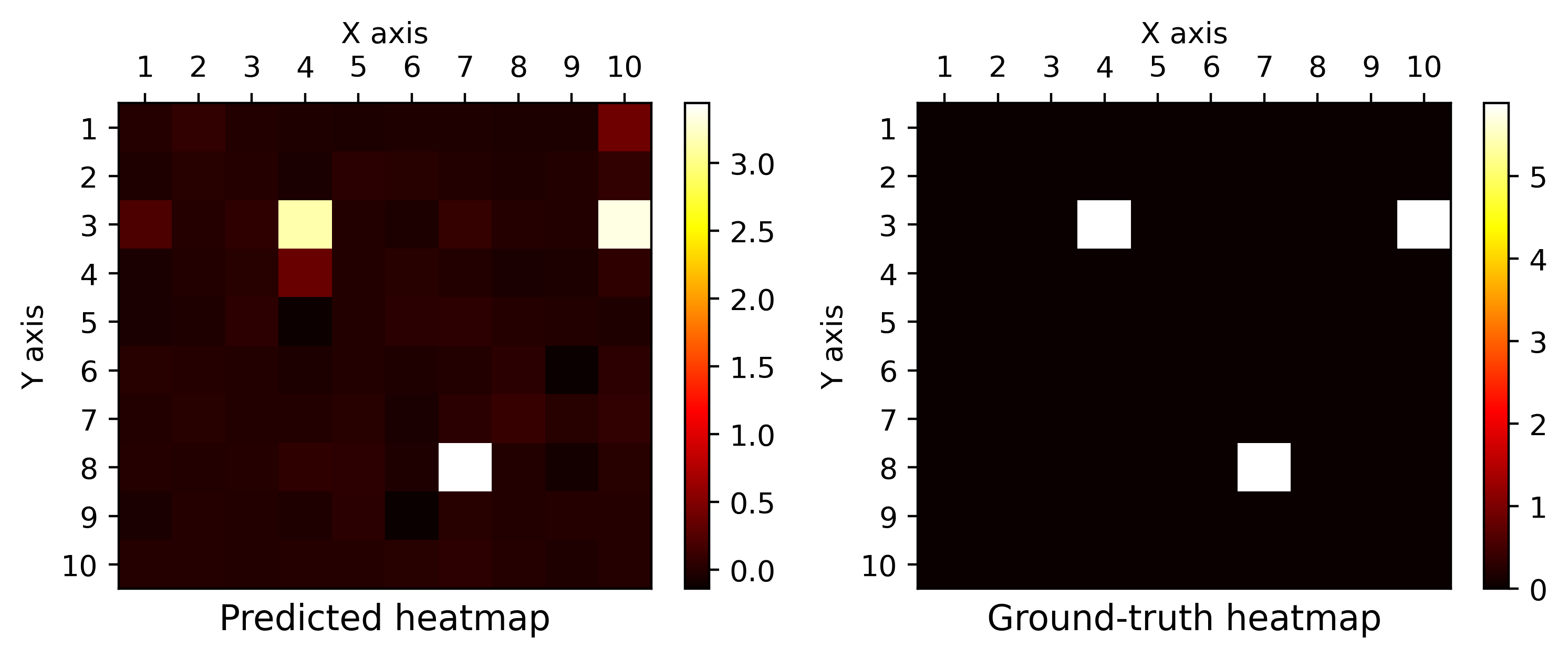}
      \caption{Good case for the triple-contact probe}
      \label{fig:good_case3}
      \vspace{-2pt}
      \end{subfigure}
      \caption{Good heatmap predictions: contact locations were predicted accurately with non-trivial force errors (in Newton).}
      \label{fig:good_samples_on_robot_parallel}
      \vspace{-5pt}
  \end{figure*}

    \begin{figure*}[!t]
    \vspace{-5pt}
      \centering
      \begin{subfigure}{0.329\linewidth}
      \includegraphics[width=1.0\linewidth]{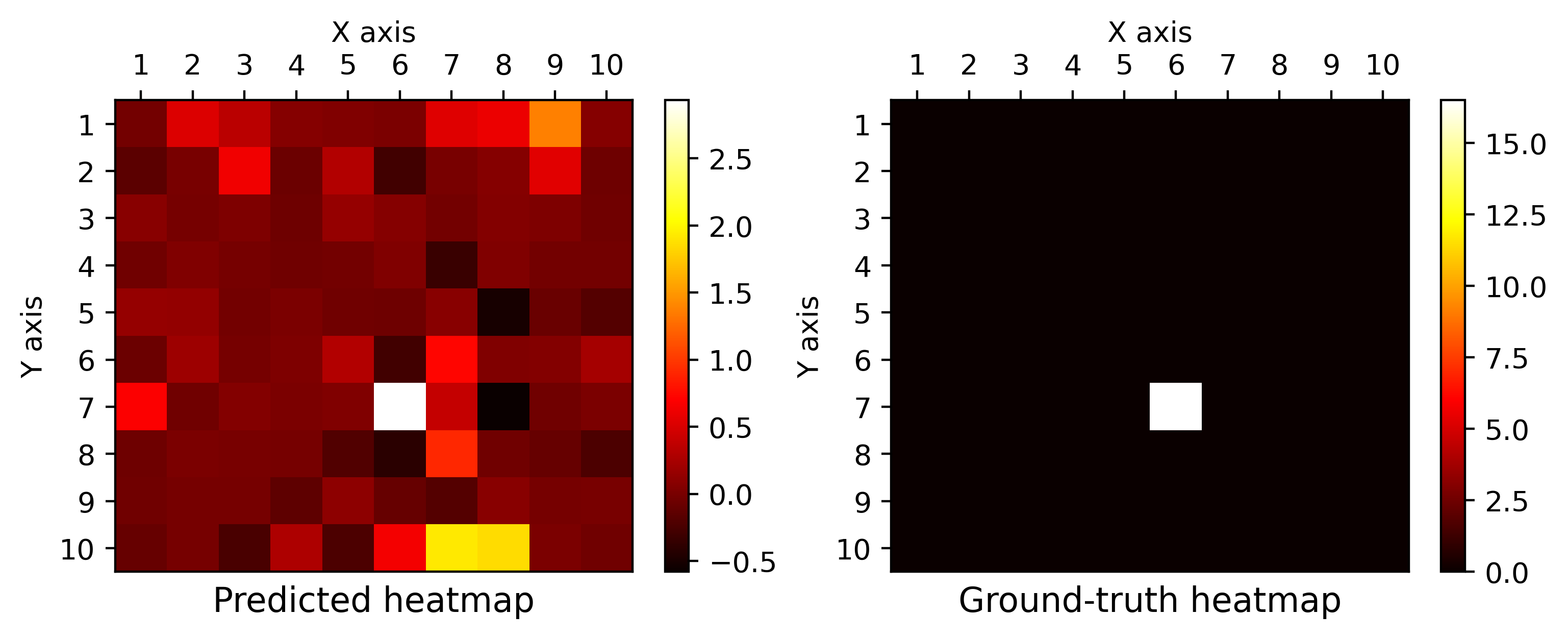}
      \caption{Poor case for the single-contact probe}
      \label{fig:poor_case1}
      \vspace{-2pt}
      \end{subfigure}
      \begin{subfigure}{0.329\linewidth}
      \includegraphics[width=1.0\linewidth]{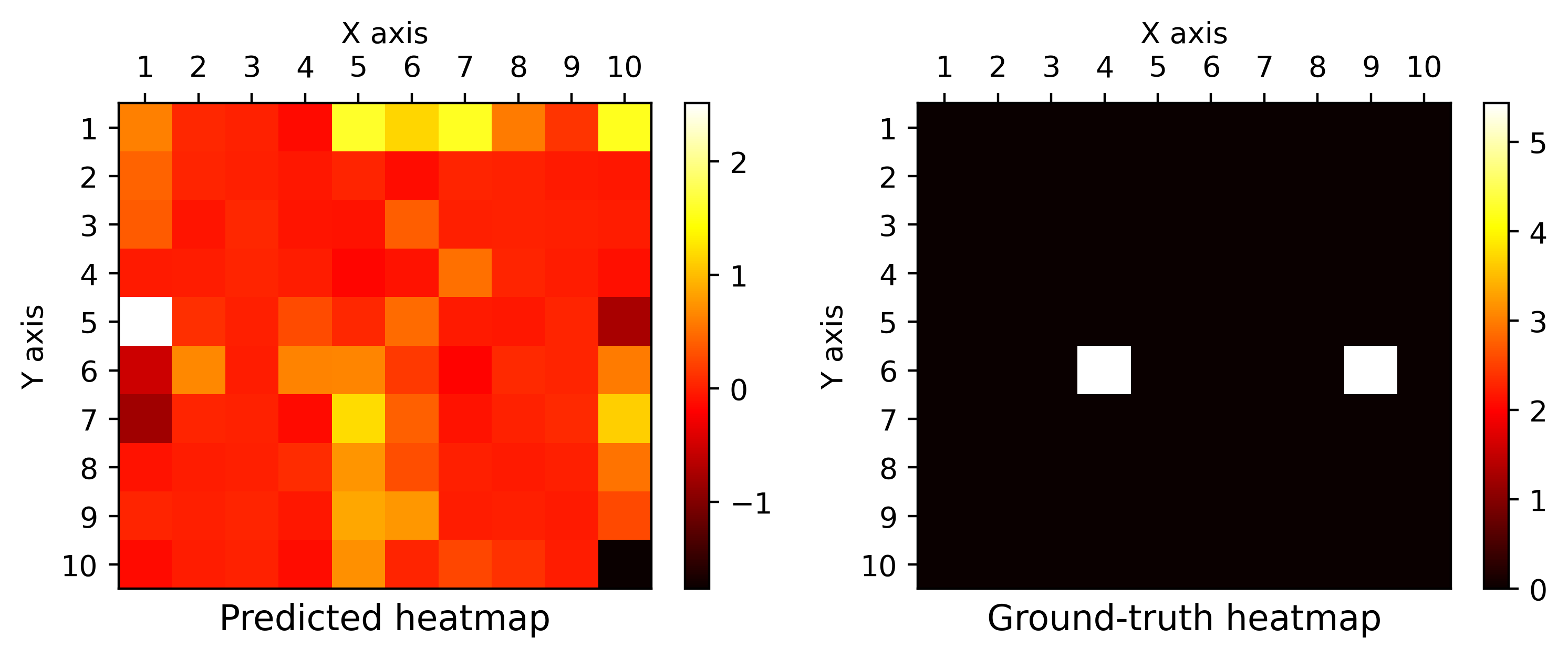}
      \caption{Poor case for the dual-contact probe}
      \label{fig:poor_case2}
      \vspace{-2pt}
      \end{subfigure}
      \begin{subfigure}{0.329\linewidth}
        \includegraphics[width=1.0\linewidth]{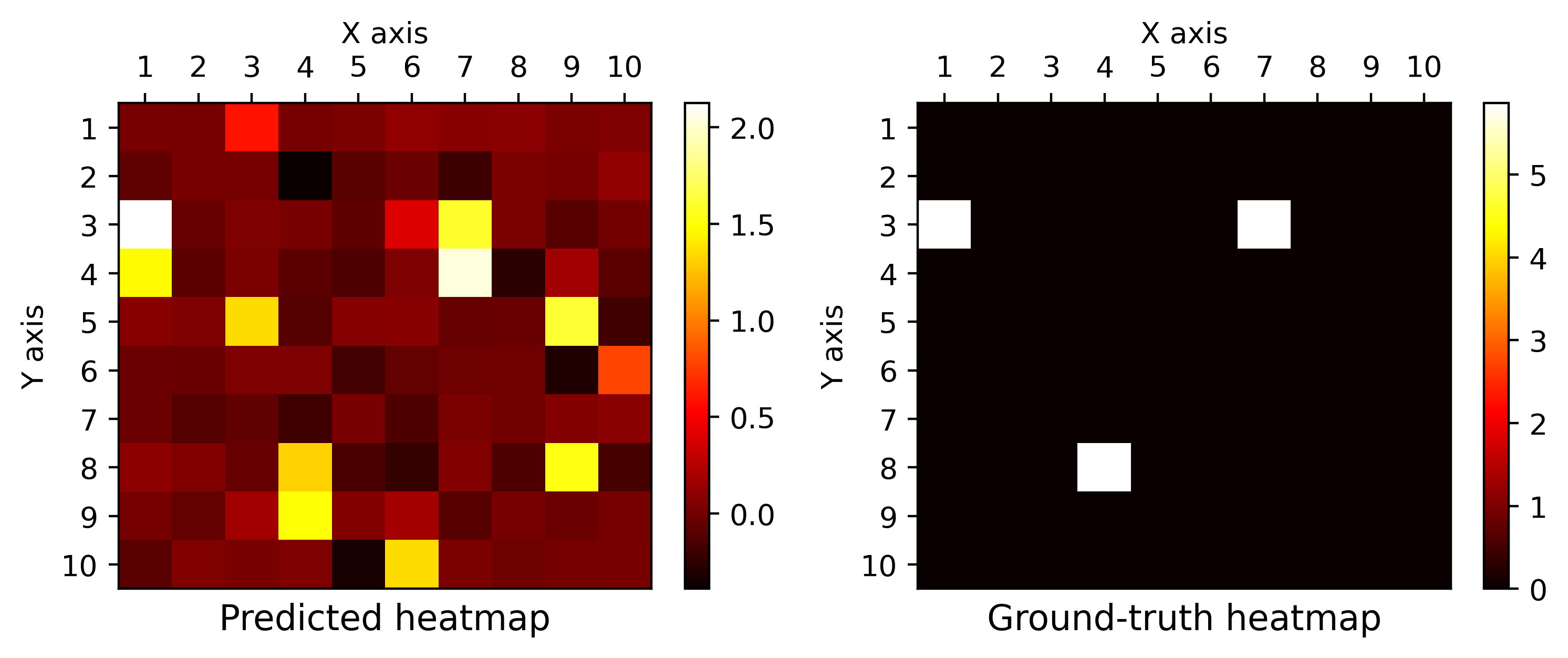}
      \caption{Poor case for the triple-contact probe}
      \label{fig:poor_case3}
      \vspace{-2pt}
      \end{subfigure}
      \caption{Poor heatmap predictions: both predicted contact locations and applied forces (in Newton) had large errors. 
      }
      \label{fig:poor_samples_on_robot_parallel}
      \vspace{-12pt}
  \end{figure*}

To evaluate how the instrumented object could perform in real applications, experiments were conducted by using the object to measure contact locations and applied forces when being touched by a parallel jaw gripper (Fig.~\ref{fig:overall}). In the experiment, the instrumented object was touched by two probes attached to two load cells (providing ground-truth normal force) installed on a Franke Emika Panda gripper.
A 10$\times$10 grid was painted on each face and two cameras were used to observe the experiments for providing ground-truth contact locations (manually labeled).
The three probes used for data collection in Section~\ref{sec:data_collection} were also used here to measure the performance with different numbers of contacts on opposite faces (\textbf{Face3} and \textbf{Face5}).
Samples were collected by touching the object with forces from 0 to around 17N (controlled by gripper displacements) in five different locations for the single-contact probe, three locations for the dual-contact probe, and two locations for the triple-contact probe. All locations were uniformly distributed on both faces to cover most regions of the object surfaces. The same models in Section~\ref{sec:general_performance} were used here.

Table~\ref{tab:real_test_parallel} lists its average performance in all metrics for 4000 samples in total. From the results, we can observe that both faces have high non-contact accuracy and location accuracy: $E_{loc} < 2.8$ (the Euclidean error of the case with two pixels' labeling error in both axes) can be considered as no error, due to the inevitable manual label errors for relatively large probes (with a radius of 5mm) and small pixels (with an edge-length of 3mm). The location labels in Section~\ref{sec:data_collection} do not have these errors as they were collected from the CNC machine rather than manual labeling.

However, the force errors and heatmap similarities were much worse than those in Section~\ref{sec:general_performance}. 
Fig.~\ref{fig:good_samples_on_robot_parallel} and Fig.~\ref{fig:poor_samples_on_robot_parallel} show how the design performed in good and poor cases, where we can see that location estimates were accurate with non-trivial force errors in good cases, but estimation errors in both contact location and force were large in poor cases. More results are shown in the video: \href{https://youtu.be/kQSZlNxYRrs}{https://youtu.be/kQSZlNxYRrs}.

One factor that could cause this problem is internal core slipping/shifting w.r.t. the outer shell between calibration and evaluation. These shifts caused as large as 1000 (37\% of its range) Hall signal differences between the calibration (i.e., training data collection in Section~\ref{sec:data_collection}) and the evaluation in this section, even for non-contact situations. Although an approach was adopted to compensate for possible Hall signal changes by using signals from non-contact situations as reference offsets for Hall signals in contact situations (which did benefit the non-contact accuracy), it remains inadequate in effectively managing the problem caused by these shifts.

Another factor is that the influence of the contacts from other faces is non-trivial, particularly when the contacts from other faces are strong, as shown in Fig.~\ref{fig:performance_vs_force_range}). This breaks the assumption of the method (Section~\ref{sec:contact_estimation}) that the Hall signal changes resulting by the contacts on other faces are negligible. From Fig.~\ref{fig:performance_vs_force_range} we can see that the performance of both contact and non-contact faces became worse when the contacts from other faces increased from 0.3 to 1.0 (normalized scale).
There might be other reasons causing the problem. More systematic studies are required in the future to seek better designs (see Section~\ref{sec:design_considerations} for detailed discussions).

\section{Design Considerations}
\label{sec:design_considerations}
The evaluation in Section~\ref{sec:eval} and Section~\ref{sec:eval_real} demonstrated the effectiveness, but also showed the problems, of the current design for assessing compliant robotic grasping. This section discusses the problems 
and potential remedies.

\paragraph{Reducing internal position variances}
The first problem is the internal core shifting w.r.t. the silicone shell between calibration and evaluation. Appropriate mechanisms need to be explored for reducing these shifts.

\paragraph{Reducing the influence from contacts on other faces}
With the current design, contacts on one face can produce magnetic field changes on other faces, therefore causing unexpected Hall signal changes and affecting the contact estimation accuracy of other faces. 
One option is swapping the magnets and Hall sensors (i.e., placing the magnets in the internal core, but the Hall sensors in the silicone shell). Then deformations on other faces will not affect the magnetic field, although there will still be the cross-coupling influence caused by the deformation indirectly produced by the contacts from other faces. 
Another option is replacing the rare earth magnets with electromagnets. Electromagnets can be activated individually for each face, therefore eliminating the magnetic field influence from other faces.

However, their fabrication becomes more challenging, as both need circuits in the silicone shell for either Hall sensors or electromagnets. The latter option could also bring in extra challenges in power supply and cooling for the electromagnets, but it can potentially open possibilities in efficient and accurate contact estimation with designed electromagnet patterns.
Other solutions include collecting training data with cross-coupling effects (which could be costly due to a large number of coupling combinations) or refining the mechanical design to reduce/avoid deformations caused by contacts from other faces.

\paragraph{Improving the accuracy in edge areas}
Results in Section~\ref{sec:loc_wise_eval} showed that the arrangement of magnets and Hall sensors is crucial for uniform performance in different locations. 
To achieve good coverage of all/most cases with as few sensors as possible, the arrangement of the magnets and Hall sensors needs more investigation. Finite element analysis could be used to help search for good designs by taking into account both contact estimation performance and instrumented object physical properties (the magnets, sensors, wires, and all other components in the object can affect its physical properties such as stiffness or compliance in different parts).

\paragraph{Balancing the spatial and temporal coverage of training data}
Studies in Section~\ref{sec:unseen_vs_seen} and Section~\ref{sec:data_amount} showed that the location coverage of training data is critical for better contact location/case estimation accuracy, while much less data (2683 samples in total, and about 19 samples for each contact location/case) can enable accurate contact force estimation. In future training/calibration data collection, a better balance can be achieved by collecting less data for each contact location/case but covering as many contact locations and numbers of contacts as possible.

\section{Conclusion}
A new concept was introduced in this paper to assess compliant robotic grasping from a first-object perspective using instrumented objects. The feasibility of the concept was demonstrated with a proof-of-concept design which made use of Hall-effect sensors and data-driven methods to interpret the signals for estimating contact location and normal force. The design generally worked well in contact estimation, although there were some design drawbacks that caused non-trivial performance degradation with simultaneous contacts from multiple faces and the internal core shifting w.r.t. the silicone shell between calibration and evaluation.

This concept and design can be extended in the future for more comprehensive measurements of compliant robotic grasping (such as the measurement of 3D contact force and more simultaneous contact points, and the estimation of deformation and shape changes). Ultimately this could lead to a robust framework for autonomously generating designs of instrumented objects with various measurement capabilities and physical properties (shapes, sizes, stiffness, etc) for benchmark purposes. Such instrumented objects could close the loop of developing/learning solutions for different compliant robotic manipulation tasks by providing timely and accurate feedback from a first-object perspective.

\section*{Acknowledgment}

The authors acknowledge continued support from the Queensland University of Technology (QUT) through the Centre for Robotics. This work was also supported by the QCR ECR/MCR scheme. Support from the GentleMAN (RCN 299757) is also greatly acknowledged.

\bibliographystyle{IEEEtran}
\bibliography{reference}

\end{document}